%% file: main.tex
\documentclass[10pt,twocolumn,letterpaper]{article}

\usepackage{cvpr}              
\usepackage{kotex}

\input{preamble}

\definecolor{cvprblue}{rgb}{0.21,0.49,0.74}
\usepackage[pagebackref,breaklinks,colorlinks,citecolor=cvprblue]{hyperref}

\title{Text-Guided Variational Image Generation\\for Industrial Anomaly Detection and Segmentation}           

\author{Mingyu Lee\textsuperscript{1,2} \,\,\,\,\,\,\,\,\,\,\,\,\,\, Jongwon Choi\textsuperscript{1}\thanks{Corresponding author.}\\
\textsuperscript{1}Dept. of Advanced Imaging, GSAIM, Chung-Ang University, Seoul, Korea\\
\textsuperscript{2}Generative AI Team, LG CNS, Seoul, Korea\\
{\tt\small mglee@vilab.cau.ac.kr, choijw@cau.ac.kr}
}

\AddToShipoutPicture*{%
     \AtTextUpperLeft{%
         \put(0,30){
           \begin{minipage}{\textwidth}
              \footnotesize
              Preprint version; final version will be available at \url{https://openaccess.thecvf.com}\\
              The IEEE / CVF Computer Vision and Pattern Recognition Conference (CVPR) (2024)\\
              Published by: IEEE \& CVF
           \end{minipage}}%
     }%
}

\begin{document}
\maketitle

\begin{abstract}
We propose a text-guided variational image generation method to address the challenge of getting clean data for anomaly detection in industrial manufacturing. Our method utilizes text information about the target object, learned from extensive text library documents, to generate non-defective data images resembling the input image. The proposed framework ensures that the generated non-defective images align with anticipated distributions derived from textual and image-based knowledge, ensuring stability and generality. Experimental results demonstrate the effectiveness of our approach, surpassing previous methods even with limited non-defective data. Our approach is validated through generalization tests across four baseline models and three distinct datasets. We present an additional analysis to enhance the effectiveness of anomaly detection models by utilizing the generated images.
\end{abstract}

\section{Introduction}
Identifying anomalous components in industrial manufacturing, a task known as anomaly detection, has been a challenging but important problem to solve.
Conventional methods for anomaly detection solve this problem by training the distribution of non-defective data in an attempt to identify defects~\cite{ref:svdd,ref:patch-svdd,ref:deep-svdd,ref:padim,ref:patchcore,ref:fewanomaly}.
Recently, it has also been proposed to train invertible functions with the images to Gaussian distribution using the probabilistic method without adaptation to the target distribution~\cite{ref:differnet,ref:cross-scale-flows,ref:cflow,ref:fastflow}.

The efficacy of anomaly detection depends on the quantity and quality of available non-defective data, as these factors directly impact the model's ability to encompass the diverse spectra of target object appearances.
Most of all, it is difficult to visually classify industrial defects since errors can vary from subtle changes as thin scratches to significant structural defects as missing components. 
Such difficulty naturally falls into the problem of out-of-distribution detection, in which a model must differentiate the samples obtained from the training data distribution from those outside of its range.
As a result, it is essential to obtain various non-defective data that effectively represent the data distribution and are distinguished from defective cases.

\begin{figure}[t]
  \centering
  \includegraphics[width=\linewidth]{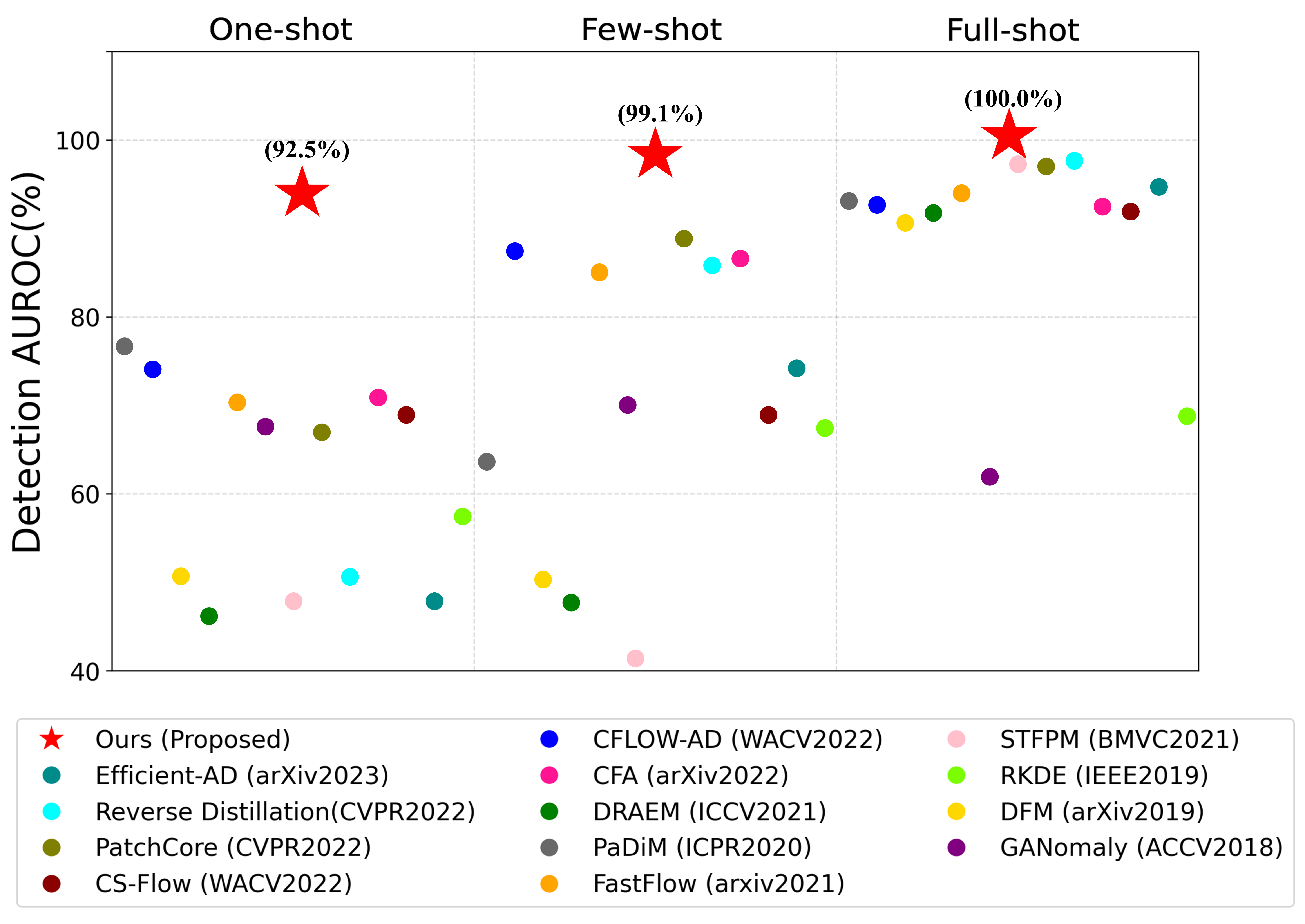}
       \vspace{-0.65cm}
  \caption{\textbf{Comparison with state-of-the-art baselines.} 
Our method generates non-defective images using a text-guided variational image generation method and utilizes the generated images as additional training data for anomaly detection. 
Ours outperforms the state-of-the-art methods across various settings, such as one-shot, few-shot (5 images), and full-shot training images. 
For comparison, we use the metal-nut class of MVTecAD dataset~\cite{ref:mvtec}.}
  \label{fig:teaser}
      \vspace{-0.49cm}
\end{figure}

However, the industrial field suffers from various data issues.
The first is the issue of imbalanced non-defective data, where the non-defective data itself can present a near-uniformity with only a few images exhibiting minor differences that qualify as acceptable defects.
The second issue is related to the sensitivity of the data, especially in appearance and level of defectiveness, as they can vary significantly due to the type of machine used and the capturing conditions.
The last one is the potential to wrongly label defective data as non-defective data by mistake during the excessive collection of non-defective data.

To overcome the issues posed by the requirement of large non-defective data, we propose a text-guided variational image generation method for industrial anomaly detection and segmentation.
To solve the problem of lack of diversity in the provided non-defective data, we extensively utilize text information about the target object learned through comprehensive text library documents to generate non-defective data images that are most similar to the input image.
In addition, it is important to recognize that the normal distribution can also be influenced by the given set of non-defective image data, with its behavior being modulated by the variance predicted by our new framework of variational image generator.
Integrating textual information and image data ensures that our proposed framework can generate several non-defective images while upholding the anticipated non-defective distribution derived from textual and image-based prior knowledge.
By utilizing a bunch of non-defective images generated from our generator, we can employ our framework in various anomaly detection methods, thereby potentially enhancing their overall performance.
We validated the proposed method through generalization tests across four baseline models and three distinct datasets.
Interestingly, as shown in Fig.~\ref{fig:teaser}, even relying on a single or a few non-defective images, our method outperforms the previous methods requiring many non-defective data.

\vspace{1mm}
We can summarize our contribution as follows:
\begin{itemize}
    \item
    To ensure generalization and robustness, we developed a variation-based image generator to predict and preserve the variance of the provided non-defective images.
    \item
    To solve the problem of lack of diversity in good product data, we developed a keyword-to-prompt generator that generates the best prompt by comparing text information about the target object, learned extensively through comprehensive text library documents, with the input image.
    \item
    To bridge the semantic gap arising from different modalities, we developed a text-guided knowledge integrator method in which latent image features are aligned with the text information of the target object.
    \item
    To validate the efficacy of our approach, we have merged our method into several state-of-the-art algorithms and tested extensively across various real-world industrial datasets, and the experimental results confirm that our framework shows impressive performance even with a single or a few non-defective images.
\end{itemize}

\section{Related Works}
\noindent\textbf{Text-based Anomaly Detection.}
Most Anomaly detection methods include representation-based methods that extract discriminative features for patches and calculate distances between distributions~\cite{ref:fastflow,ref:cflow} and reconstruction-based methods~\cite{ref:draem,ref:patchcore} using generative adversarial networks.
With anomaly detection advances, benchmark dataset performance in full shot settings has become saturated due to ~\cite{ref:mvtec} state-of-the-art research~\cite{ref:efficientad,ref:memseg,ref:PNI}.
However, recent studies employ different settings and methods, such as text-based anomaly detection.
For example, WinCLIP~\cite{ref:winclip} generates pre-text prompts for the input image by mixing state words and text candidates. 
Then, it extracts multi-scale features of the input image through a binary mask and calculates a similarity score between the generated prompts to detect anomalies.

In contrast, our method focuses on generating non-defective images based on the similarity of the input image and the text-based prior knowledge. It improves performance by feeding the newly generated images of different attributes into the general anomaly detection methods.

\noindent\textbf{Text-to-Image Generation.}
Recently, text-to-image generation models~\cite{ref:clip,ref:cliphome} are widely studied, which predict or generate the next pixel values based on the given text and all previous predictions of pixel values.
For example, DALL-E~\cite{ref:dall-e} takes text input from the GPT series model to predict the next pixels, which are then added to the initial image to be placed as the subsequent input along with the text. By repeating the process multiple times, DALL-E can generate an entire image.
GAN~\cite{ref:gan} is known for high-quality and realistic image generation without a prompt, and Contrastive Language-Image Pre-training (CLIP) is another neural network that can determine how well a caption or prompt matches an image.
Through VQGAN-CLIP~\cite{ref:vqgan-clip}, where VQGAN~\cite{ref:vqgan} creates an image, CLIP~\cite{ref:clip} checks whether the generated image from VQGAN matches the text.
When the process is repeated, the created image gradually resembles the input text.

However, the previous text-guided image generation methods ignore the importance of the representation variance of the given image data, which is critical for generalizing the non-defective image data in anomaly detection tasks.
In contrast to the previous text-based image generation methods, we aim to predict the variance of non-defective image distribution, which is utilized to avoid generating defective images while enlarging the variety of appearances in non-defective images.

\section{Preliminary Analysis}
\begin{figure}[t]
\centering
    \includegraphics[width=\linewidth]{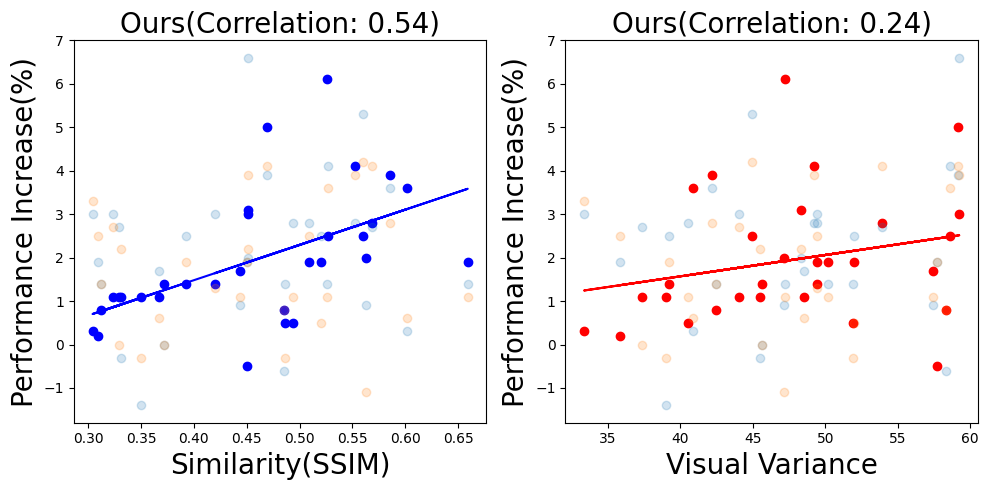}
    \vspace{-0.6cm}
\caption{
\textbf{Correlation of hypothesis.}
We repeat the tests with different images to confirm our hypothesis.
Performance enhancement is strongly connected with the similarity between the generated and original images, and the visual variance of created images also improves performance.
}
  \label{fig:Correlation experiment}
\end{figure}

\noindent\textbf{Hypothesis.} Data issues related to sensitivity, especially in appearance and level of defectiveness, can vary significantly due to the type of machine used and the capturing conditions.
Thus, the similarity between the generated and original images is highly correlated with performance enhancement, and the visual variance of generated images also contributes to performance increases.
In Fig.~\ref{fig:Correlation experiment}, the similarity between the generated and original images is highly correlated with the performance enhancement, and the variance of generated images also contributes to the performance increase. 
These results empirically validate our hypothesis that the generated images should be similar to the appearance of the provided non-defective images while preserving their visual variance. 
Furthermore, we compared the correlation between the original image and the naive prompts (0.39), just words (0.35), and ours(0.54), which confirms the necessity of well-designed prompts in our hypothesis.

\noindent\textbf{Preliminary Analysis.}
We repeat the preliminary experiments using various images with a simple scenario to verify our hypothesis in general.
In the scenario, we compare the performance between the baseline anomaly detection model and the same model using one additional training image searched on the web or generated by an image generator.
We consider a single \textit{hazelnut} image from the MVTecAD dataset~\cite{ref:mvtec}, utilizing Patchcore~\cite{ref:patchcore} for the baseline anomaly detection model.

\begin{figure}[t]
\centering
    \includegraphics[width=\linewidth]{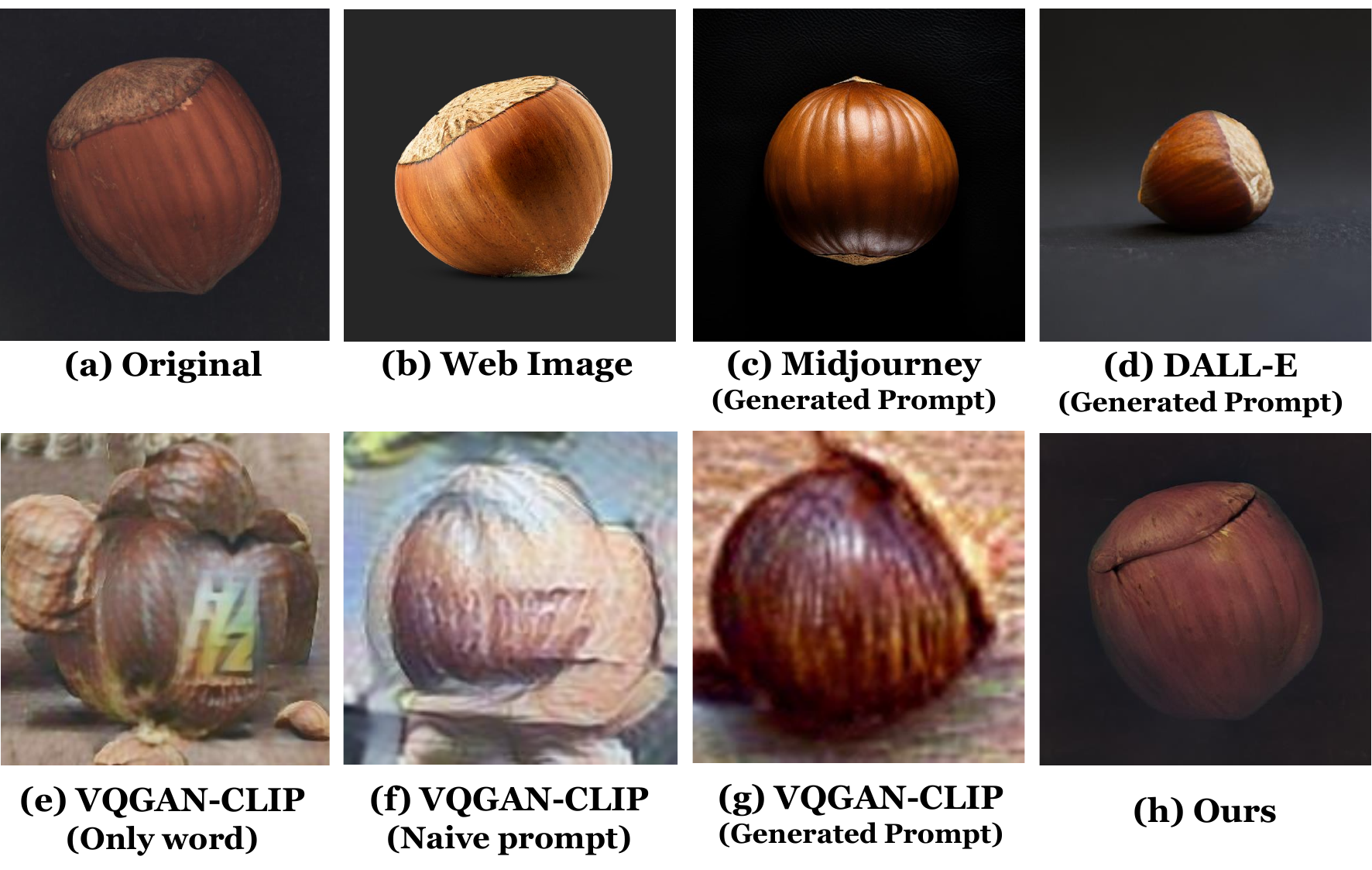}
    \vspace{-0.6cm}
  \caption{\textbf{Generated images in the preliminary experiment.} 
(a) The original image of \textit{hazelnut} in MVtecAD dataset.
(b) The image retrieved by a keyword of `hazelnut' from the web.
(c),(d) The images generated using Midjourney and DALL-E, respectively, using a captioning of the original image as a prompt.
(e),(f),(g) The images generated using the VQGAN-CLIP model based on `hazelnut', `A photo of a hazelnut', and the captioning of the original image, respectively.
(h) The image generated by our method.
}
  \label{fig:Visualize of preliminary experiment}
\end{figure}

We build various types of generated images :\\
\textit{Original Image}, \textit{Web Image}~\cite{ref:GoogleImages}, 
\textit{Midjourney}~\cite{ref:MidJourney}, and \textit{DALL-E}~\cite{ref:dall-e},
\textit{VQGAN-CLIP(only Word)}, and \textit{VQGAN-CLIP(Na\"ive Prompt)}.
First, \textit{Original Image} utilizes the given single original image to train the anomaly detection model, while \textit{Web Image} adds one additional image retrieved from the web by using a keyword of ``Hazelnut.''
For the text-based image generator, we first generated the text captioning from the original hazelnut image by using GPT-4, and the result was:
\textit{``This image displays a single hazelnut with a textured, fibrous cap on top. The nut itself has a rich, one image was created using warm brown color with visible stripes and markings that suggest a natural origin. It rests against a dark background which serves to highlight the nut's detailed texture and organic shape.''}.
\textit{Only Word} and \textit{Na\"ive Prompt} utilize the prompts of ``Hazelnut" and ``A photo of a hazelnut", respectively.
The three types of prompts were fed into \textit{Midjourney}, \textit{DALL-E}, and \textit{VQGAN-CLIP} to generate the additional training image.
Fig.~\ref{fig:Visualize of preliminary experiment} shows the additional training image from various generation schemes, presenting that the results highly depend on the types of image generation models and the prompts.

\begin{table}[t]
\caption{
\textbf{Preliminary experiment using Patchcore.~\cite{ref:patchcore}}
The results the need for a well-designed generation model to boost anomaly detection performance by adding extra training images.} 
  \centering
\centering
    \vspace{-0.3cm}
\resizebox{\linewidth}{!}{
\begin{tabular}{l|c|c}
    \noalign{\smallskip}\noalign{\smallskip}\hline\hline
                Method & Detection & Segmentation \\
    \hline
    \hline 
    \text {Original Image}  &\textbf{88.3} &\textbf{94.3} \\ 
    \text {Web Image}                   &84.8 &93.1 \\ 
    \text {Midjourney (by Generated Prompt)}                  &82.7 &93.3 \\  
    \text {DALL-E (by Generated Prompt)}                      &85.7 &91.9 \\  
    \hline
    \text {VQGAN-CLIP (by Only word)}        &64.8 &86.8 \\  
    \text {VQGAN-CLIP (by Na\"ive Prompt)}   &71.9 &88.2 \\  
    \text {VQGAN-CLIP (by Generated Prompt)}   &81.4 &91.7 \\  
    \hline
    \text {\bf Ours }                        &\textbf{93.3} &\textbf{94.6} \\  
    \hline
    \hline
\end{tabular}
}
\label{tb:Comparison of preliminary experiment}
\vspace{-0.5cm}
\end{table}

The results of preliminary experiments are represented in Table~\ref{tb:Comparison of preliminary experiment}, showing that performance is affected depending on the generated images.
Interestingly, preliminary results show that good image quality does not necessarily help improve performance as presented in the low performance of DALL-E.
Meanwhile, given the performance of web images is relatively high, it can be assumed that performance can be improved if elements such as outlines and angles of the original image are preserved in the additional images.

\begin{figure*}[t]
\centering
    \includegraphics[width=\linewidth]{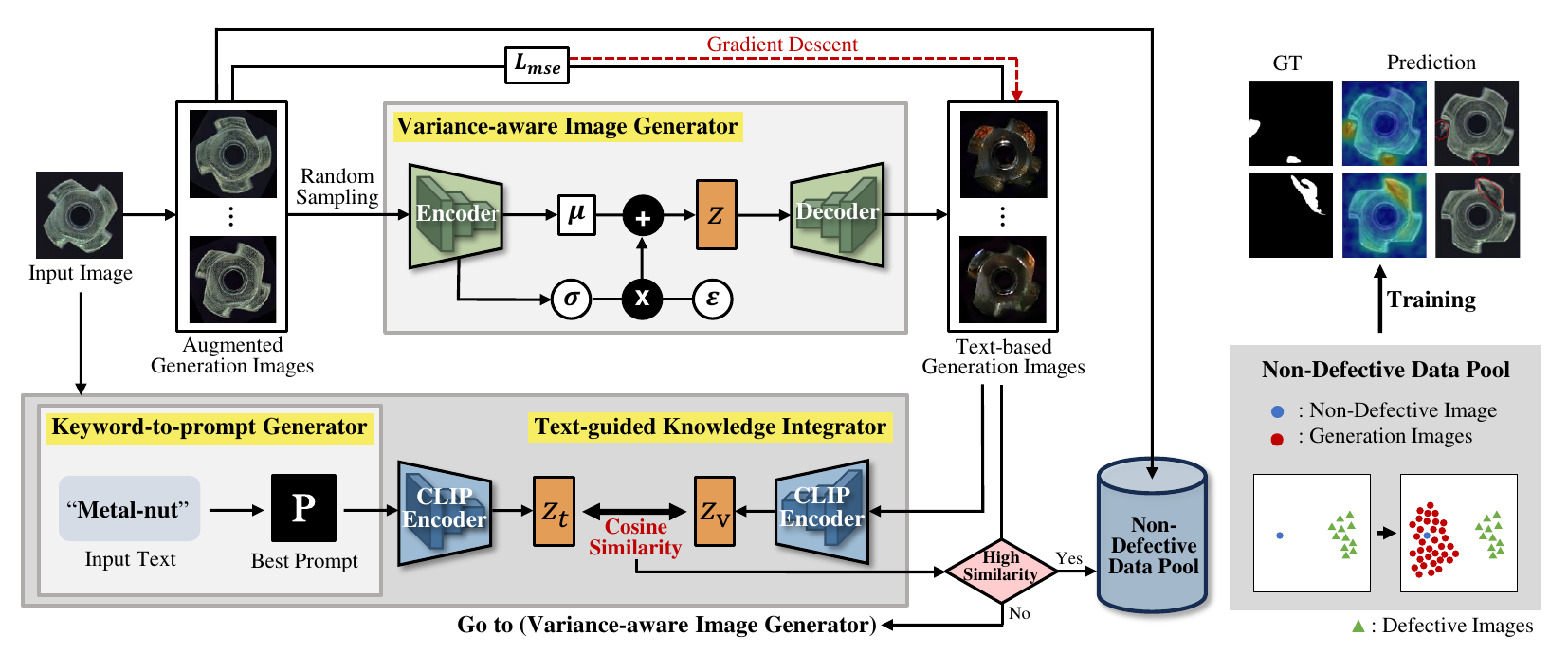}
\vspace{-0.8cm}
  \caption{  
  \textbf{Overview of our framework.} Our framework comprises a keyword-to-prompt generator, a variance-aware image generator, and a text-guided knowledge integrator. 
The keyword-to-prompts generator creates prompts from key input words and selects the best one that matches an input image. 
A variance-aware image generator creates non-defective images, encoding their visual features into a normal distribution to maintain variance. 
Our process updates through iteration, and a text-guided knowledge integrator selects the optimal images by comparing the similarity of their latent distribution to the text prompts.
}
  \label{fig:framework}
\vspace{-0.1cm}
\end{figure*}

Therefore, while simply generated images cannot be utilized for anomaly detection models, we need to consider several additional factors for generated images to train non-defective distribution effectively.
First, generated images should be similar to the appearance of the provided non-defective images while preserving their visual variance.
Second, it is important to find the best prompt to generate visually well-structured images.
Lastly, based on the above two information contents, we should create images with a small semantic gap even when an insufficient number of non-defective images is given.

\section{Methodology}
As shown in Fig.~\ref{fig:framework}, our framework comprises the keyword-to-prompt generator, variance-aware image generator, and text-guided knowledge integrator.
The keyword-to-prompt generator produces a group of prompts consisting of different word combinations based on the essential words presented in the input text, then selects the best prompt that contains comparable information to the input image among the generated ones.
The variance-aware image generator generates a set of non-defective images, encoding the visual features of the non-defective images into the normal distribution to preserve their variance. 
The generation process is iterated with the generator update, and the text-guided knowledge integrator determines the optimal set of generated images by assessing the latent distribution similarity between the image set and the text prompts generated by the keyword-to-prompt generator.
Finally, the optimal set of generated images is utilized as the additional training set for the baseline anomaly detection model.

\subsection{Generation Modules}
The image generation module comprises a keyword-to-prompt generator and a variance-aware image generator.
The keyword-to-prompt generator integrates the target object name $W^o$ with the set of predefined status words ($\{W_{1}, W_{2}, \ldots, W_T\}$) to generate multiple candidate prompts ($\{S_1, S_2, \ldots, S_T\}$).
Then, the keyword-to-prompt generator selects the best prompt from the candidate prompts according to their latent distances with the latent feature of the original images $I^o$.
Then, we convert the input image $\{I^o\}$ into multiple augmented images ${I_1, I_2, \ldots, I_N}$, which are fed into the variance-aware image generator, which encodes the image $I_i$ into the corresponding latent distribution $\mathbf{z}_i$.
Finally, we decode $\mathbf{z}_i$ by sampling to generate a new multiple image set $\{I^+_1, I^+_2, \ldots, I^+_M\}$.
\vspace{-0.2cm}

\subsubsection{Keyword-to-Prompt Generator}
\vspace{-0.2cm}
The keyword-to-prompt generator produces the most appropriate prompt $P$ based on the given object name $W^o$ and the original training images $I^o$.
With WordNet~\cite{ref:wordnet}, we construct a set of $T$-different words $\{W_1,W_2,\ldots,W_T\}$ to obtain the candidate prompts $(S_1, S_2, \ldots, S_T)$ where $S_t$ is ``a $\{W^o\}$ with $\{W_t\}$'' replaced by a corresponding word of $\{W_1,W_2,\ldots,W_T\}$.
Using WordNet~\cite{ref:wordnet}, we can collect a large database containing synonyms, hypernyms, hyponyms, and part-whole relationships with high semantic relevance of $W^o$.

We find the optimal prompts from the candidate prompts of $(S_1, S_2, \ldots, S_T)$ with two stages: distance-based outlier removal and embedding similarity estimation.
First, we remove the outlier prompts by using the L2 distance between the latent features of the candidate prompt and the original image.
We define the positive prompt set where the outlier prompts were removed by $S^p$, which is determined by $S^p={S_j|d(S_j, I^o) > 0.5}$, where $d(S_i, I^o) = ||f(I^o) - G(S_i)||_2$ and $f(I)$ and $G(S)$ are the image embedding vector of the image $I$ and text embedding vector of the prompt $S$, respectively.
We utilized CLIP encoders~\cite{ref:clip} to obtain the image and text embedding vectors.
Then, from the positive prompt set $S^p$, we determine the best prompt $P$ by using the cosine similarity as follows:
\begin{equation}
P = \arg\max_{S_j\in S^p} \cos \big(f(I^o), G(S_j)\big),
\end{equation}
where $\cos \big(f(I^o), G(S_j)\big)$ means the cosine similarity between the two embedding vectors of $f(I^o)$ and $G(S_j)$.
\vspace{-0.2cm}

\subsubsection{Variance-aware Image Generator}
\vspace{-0.2cm}
As a baseline architecture of the variance-aware image generator, we employ the VQGAN model~\cite{ref:vqgan}.
This model integrates a vector quantization process, which maps the continuous latent space to a discrete codebook.
Due to this characteristic, the encoder's output specifically follows the approximate posterior distribution of the latent vector, which enables to effectively reconstruct the given input image via the decoder.
The vector quantization process plays a pivotal role, as it effectively represents the image within the latent space by mapping continuous representations to a fixed codebook.
Additionally, the VQGAN model ensures that the distribution of the latent variable closely approximates the standard normal distribution.

The architecture of VQGAN can be represented as:
\begin{equation}
\begin{aligned}
I' &= p^o\big(E(I)\big) \\
E(I) &= [e\big(q_1(I)\big), e\big(q_2(I)\big), \ldots, e\big(q_D(I)\big)]\\
e\big(q_d(I)\big) &= \arg\min_{\mathbf{e}_k\in\{\mathbf{e}_1,...,\mathbf{e}_K\}} ||q_d(I) - \mathbf{e}_k ||_2,
\end{aligned}
\end{equation}
where $p^o\big(E(I)\big)$ is a decoder to generate an entire image by integrating the patches each generated from the $d$-th latent vector sampled from a normal distribution of $\mathcal{N}\{e\big(q_d(I)\big), 1\}$, $\{\mathbf{e}_1, \mathbf{e}_2, \ldots, \mathbf{e}_K\}$ are the codebook vectors embedded in VQGAN, and $K$ is the codebook size.

However, existing VQGANs uniformly utilize the variance $\sigma$ of the latent vector by 1, ignoring the possible diversity in the patch-wise appearance of the target object.
Especially, the appearance diversity must be considered for anomaly detection models where a trained distribution of non-defective images determines defects.
To address the issue, we extend the VQGAN architecture to predict the variance of latent variables.
The variance-aware image generator can be represented as:
\begin{equation}
I^+_m = p\Big( \mathbb{E}_{\forall i}[E(I_i)], \Sigma_{\forall i}\big(E(I_i)\big) \Big),
\end{equation}
where $p(E, \Sigma)$ represents the extended decoder of VQGAN sampling $d$-th patch's latent vector from the distribution of $\mathcal{N}\{E_d, \Sigma_d\}$, $E_d$ and $\Sigma_d$ represent the $d$-th column of $E$ and $\Sigma$, respectively, and $\Sigma_{\forall i}\big(E(I_i)\big)$ is a function estimating the column-wise variance vectors of $E(I_i)$ for all possible $i$.
We should remind that $I^+_m$ and $I_i$ are the $m$-th generated image and the $i$-th augmented one, respectively.
Thus, our variance-aware image generator extracts the latent variables according to the variance value estimated from the given image and the target patch.
Due to the sampling process in the VQGAN, the generated images $I^+_m$ differ at each sampling iteration $m$.

\subsection{Text-guided Knowledge Integrator}
In this procedure, we generate the non-defective images aligned well with the best prompt $P$, adding them to the non-defective data pool for the anomaly detection model.
From $P$ selected from the keyword-to-prompt generator, we extract a textual clip feature $z_t$ through a clip text encoder as $z_t=G(P)$.
At the same time, we generate the images $\{I^+_1,I^+_2,\ldots,I^+_M\}$ by using the variance-aware image generator, and the visual clip feature is estimated by averaging multiple image features as:
\begin{equation}
z_v = \mathbb{E}_{\forall m} \big[f(I^+_m)\big].
\end{equation}

The text-guided knowledge integrator repeats the generation of the image set, scoring the image set by the cosine similarity between the image feature expectation $z_v$ and the best prompt feature $z_t$.
Here, we define the $l$-th generated image set as $\{I^{l+}_1,I^{l+}_2,\ldots,I^{l+}_M\}$, and its corresponding visual clip feature as $z_v^l$.
Based on the cosine similarity between $z_v^l$ and $z_t$, we select the best set of generated images, which are used to train the following anomaly detection model.
We define the best set of generated images as:
\begin{equation}
\begin{aligned}
\{\widehat{I}^{+}_1,\widehat{I}^{+}_2,\ldots,\widehat{I}^{+}_M\} &\equiv \{{I}^{\alpha+}_1,{I}^{\alpha+}_2,\ldots,{I}^{\alpha+}_M\},\\
\alpha &= \arg\max_{l\in\{1,\ldots,L\}} \cos(z_t, z_v^l),
\end{aligned}
\end{equation}
where $L$ is the number of iteration to generate the image.

\begin{figure*}[t]
\centering
    \includegraphics[width=\linewidth]{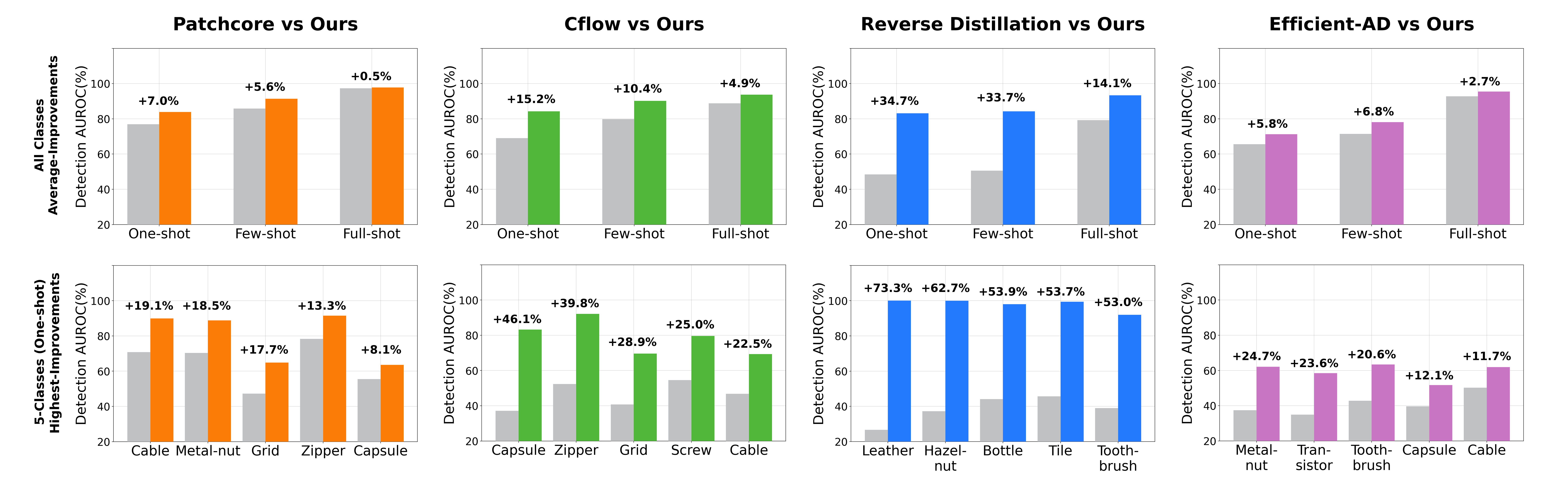}
  \caption{\textbf{Generalization test for anomaly detection in MVTecAD dataset.}
  The first row shows the average improving score across different baselines and varying numbers of non-defective images. 
  The second rows present the average score for the highest-improving five classes.}
  \label{fig:Generalization test of Anomaly detection on MVTec}
\end{figure*}

\begin{figure}[t]
\centering
    \includegraphics[width=\linewidth]{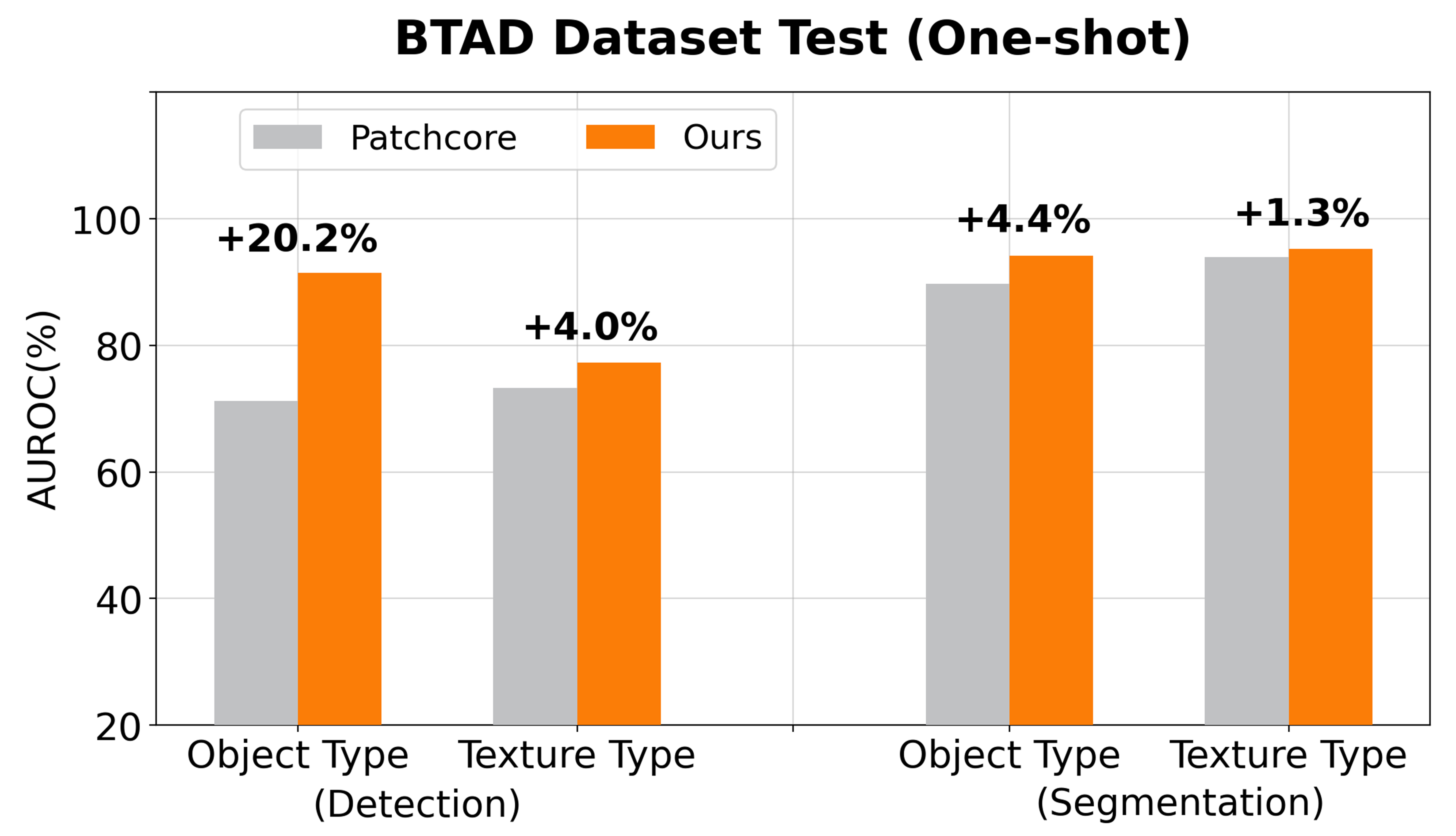}
    \includegraphics[width=\linewidth]{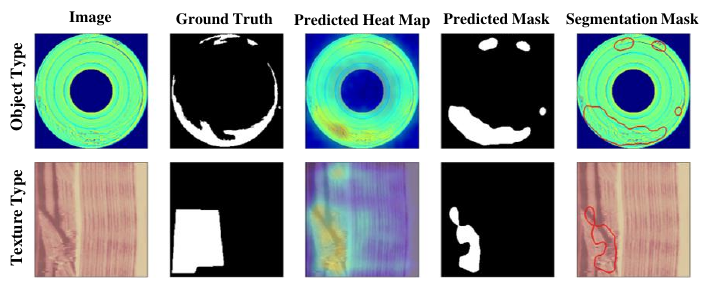} 
  \caption{\textbf{Generalization test in BTAD dataset.} 
  We have conducted experiments on detection and segmentation performance on both object and texture types of BTAD dataset~\cite{ref:btad} using  Patchcore as baseline in one-shot setting. 
  Our method improves the overall performance, particularly for the object type.}   
  \label{fig:Generalization test of Anomaly segmentation on BTAD}
\end{figure}

\begin{figure}[t]
\centering
    \includegraphics[width=\linewidth]{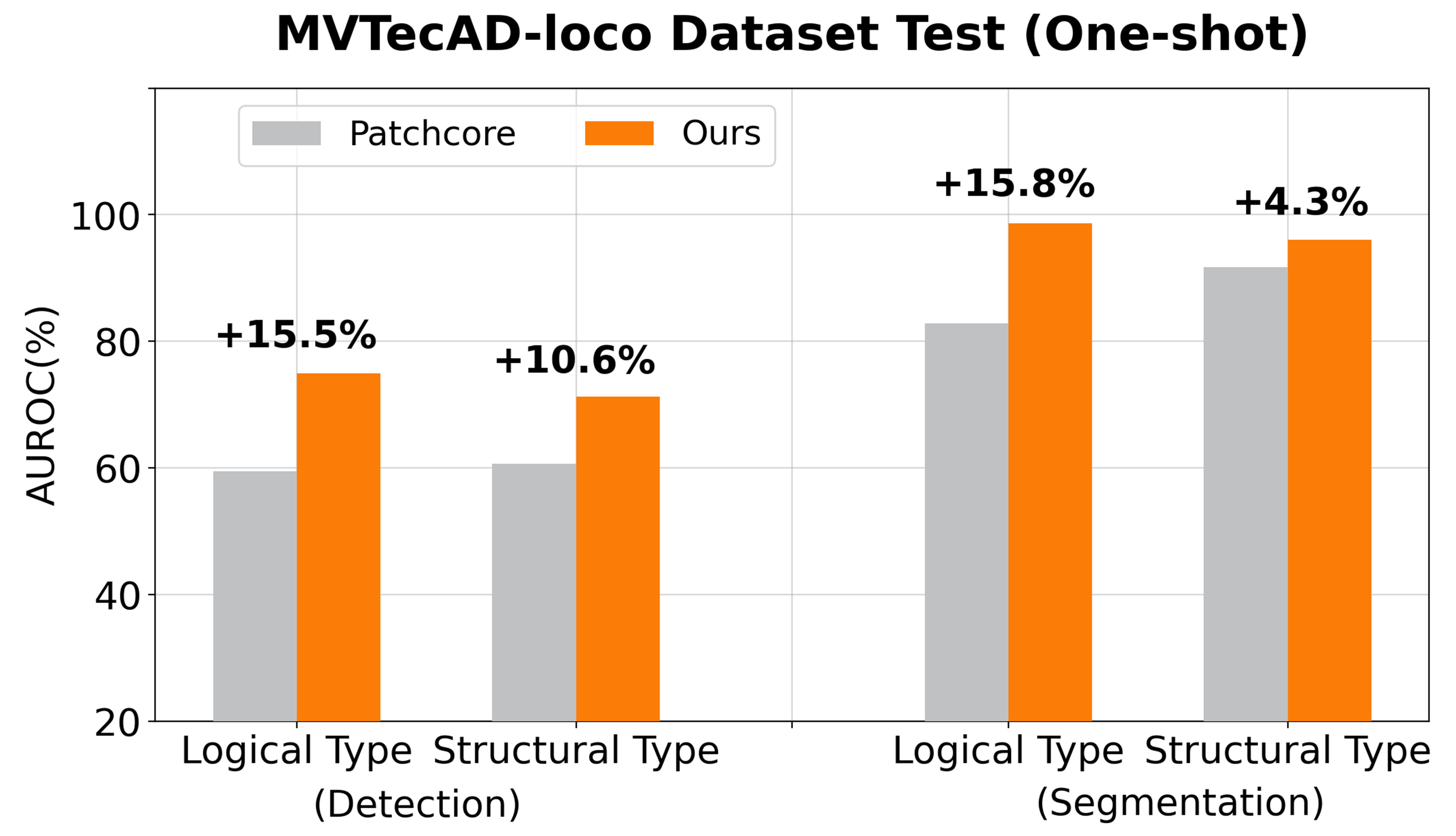}
    \includegraphics[width=\linewidth]{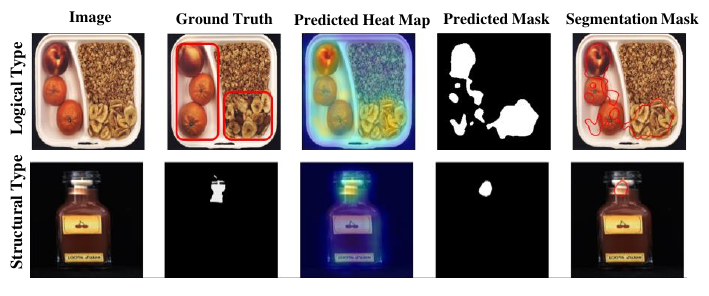}
  \caption{\textbf{Generalization test in MVTec-LOCO AD dataset.}
  We have conducted experiments on detection and segmentation performance in MVTec-LOCO AD dataset~\cite{ref:mvtec_loco}
  using Patchcore as baseline in one-shot setting.
  Our method improves the performance for overall cases.}
  \label{fig:Generalization test of Anomaly segmentation on MVTecAD-loco}
\end{figure}

At every iteration to generate the new set of images, we simultaneously update the variance-aware image generator. 
This update is necessary for two purposes: first, it is necessary for the image generator to generate images similar to the input of non-defective images, and second, we can enhance the variety of generated image sets. 
We update all parameters of the variance-aware image generator, including the encoder, decoder, and codebook vector parameters.
In addition to the original loss of VQGAN, we also utilize the Mean Squared Error (MSE) loss to generate images similar to the input images.
The MSE loss is derived as follows:
\begin{equation}
\mathcal{L}_{mse}(I^o, \{I^+_1,\ldots,I^+_M\}) = \frac{1}{n}\sum_{m=1}^{M} ||I^o - I^+_m||_2^2.
\end{equation}
Consequently, the total training loss can be represented as follows:
\begin{equation}\small
\mathcal{L} = \mathcal{L}_{mse}(I^o, \{I^+_1,\ldots,I^+_M\}) + \lambda\mathcal{L}_{vq}\big(\{I_1,\ldots,I_N\}\big),
\end{equation}\normalsize
where $\mathcal{L}_{vq}\big(\{I_1,\ldots,I_N\}\big)$ is the conventional loss for VQGAN~\cite{ref:vqgan} with the input image set of $\{I_1,\ldots,I_N\}$ and $\lambda$ is a user-defined hyperparameter. We utilize the Adam optimizer to update the estimated gradients across the variance-aware image generator.

\section{Experimental Result}

\begin{figure}[t]
\centering
    \includegraphics[width=.47\linewidth]{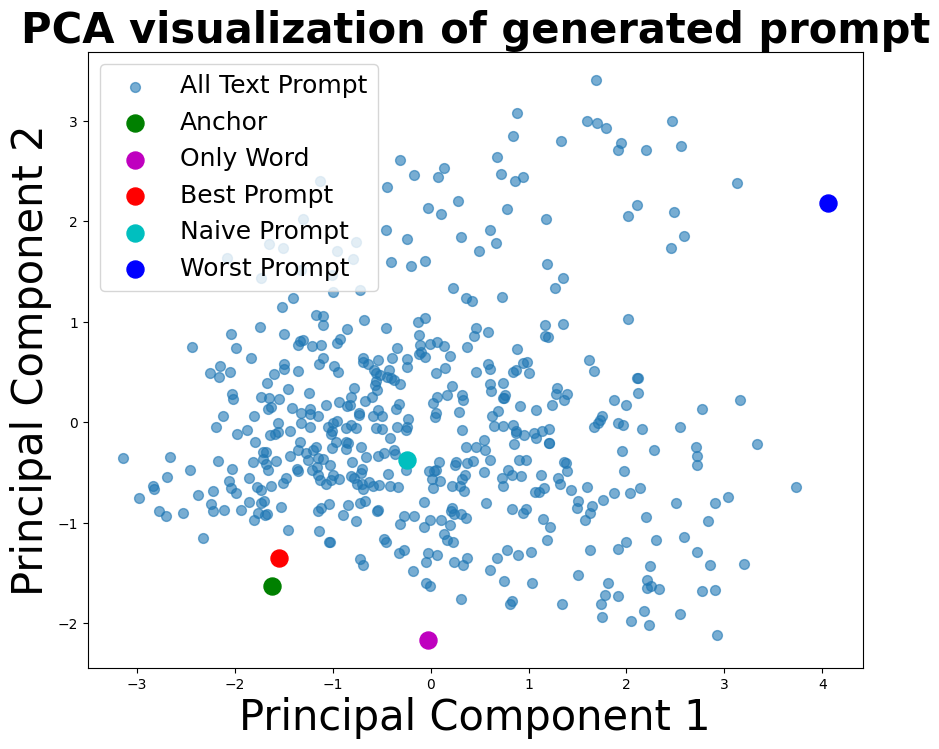}
    \includegraphics[width=.51\linewidth]{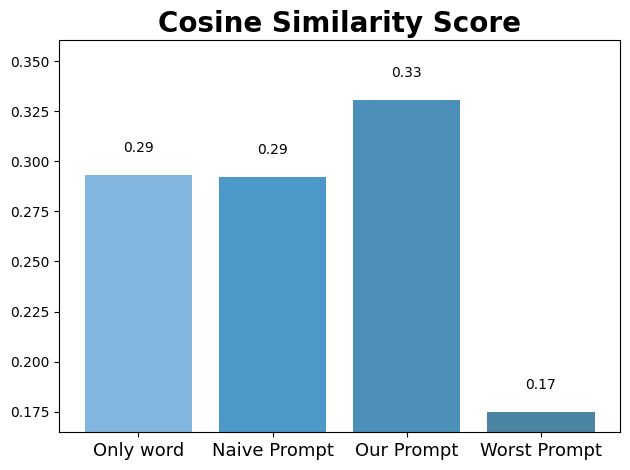}
    \includegraphics[width=\linewidth]{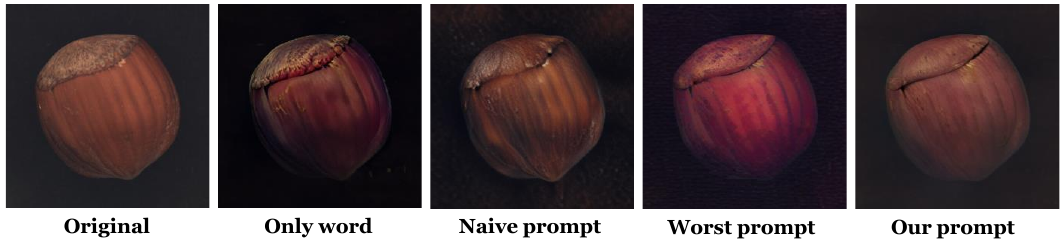}
\caption{
\textbf{(Top-Left) Similarity comparison between input text and each prompt,
(Top-Right) Similarity comparison between input image and each prompt,
(Bottom) Qualitative results of generated images.}
The prompt form by text generator is used as follows: "hazelnut" for Only word, "a photo of a hazelnut" for Na\"ive prompt, "a {\{hazelnut}\} with {\{cobnut}\}" for Our prompt, and "a {\{hazelnut}\} with {\{decantherous}\}" for Worst prompt}
  \label{fig:The comparison analysis based Cosine similarity}
\end{figure}

\subsection{Quantitative Results}
We have conducted a comparative analysis of the generalization performance of our newly designed framework for one-shot, few-shot (5 images), and full-shot (entire training images) compared to the baselines.
Firstly, the integrated average performance of the classes is derived by applying the state-of-the-art model~\cite{ref:efficientad} and the latest baselines~\cite{ref:patchcore,ref:cflow,ref:Reverse_Distillation} to the MVTecAD dataset~\cite{ref:mvtec}. 
Also, we have analyzed the five classes with the highest improvement to show the strengths of our method.
Secondly, we have conducted experiments using the BTAD dataset~\cite{ref:btad} consisting of two types of objects and textures to find the object types in which our model performs well.
Finally, we have validated the model's performance by assigning tasks with complicated object types of the MVTec-LOCO AD dataset~\cite{ref:mvtec_loco}.

In Fig.~\ref{fig:Generalization test of Anomaly detection on MVTec}, the experimental results on generalized detection using MVTecAD dataset indicate enhanced performance across all baselines, thus verifying our model's ability to generalize.
The results show an average increase of 15.7\% in one-shot, 14.1\% in few-shot, and 5.6\% in full-shot, confirming that our framework can successfully achieve impressive performance even with a few training images.
Additionally, we have analyzed the performance of the five highest improving classes in one-shot tasks, as shown in the second rows of Fig.~\ref{fig:Generalization test of Anomaly detection on MVTec}.
The results show that object types, such as the cable and metal-nut classes from the Patchcore model, show impressive improvements of 19.1\% and 18.5\%, respectively.
Other baselines show a similar trend in object types, and the reverse distillation model exhibits a significant increase of 73.3\% even in texture types such as leather.

As shown in Fig.~\ref{fig:Generalization test of Anomaly segmentation on BTAD}, we have conducted additional experiments on the BTAD dataset to revalidate which defects our model works well. 
As in the graph of Fig.~\ref{fig:Generalization test of Anomaly segmentation on BTAD} (Top), the detection AUROC shows improvements of 20.2\% for object type (1st class) and 4.0\% for texture type (2nd class), compared to baseline. 
Similarly, the segmentation AUROC has also shown more significant improvements for the object type, with 4.4\% for object types and 1.3\% for texture types.
In the qualitative results of Fig.~\ref{fig:Generalization test of Anomaly segmentation on BTAD} (Bottom), the object type has resulted in better outcomes for locating the anomaly segments close to the ground truth.
As a result, this experiment confirms that our model is more robust on the object types.

\begin{table}[t]
\caption{
\textbf{Evaluation of image quality.}
The image generated by our prompt show the highest diversity and quality based on Inception Score (IS), Structural Similarity Index (SSIM), Peak Signal to Noise Ratio (PSNR), and Learned Perceptual Image Patch Similarity (LPIPS).} 
  \centering
\centering
\resizebox{\linewidth}{!}{
\begin{tabular}{l|c|c|c|c|c}
    \noalign{\smallskip}\noalign{\smallskip}\hline\hline
            Method &Prompt & IS($\uparrow$) & SSIM($\uparrow$) & PSNR($\uparrow$) & LPIPS($\downarrow$) \\
    \hline
    \hline   
    \text Original      &N/A                                        &1.00  &1.00 &- &0.00 \\ 
    \text Only word     &``a hazelnut''                          &1.62  &0.50 &19.01 &984.15 \\
    \text Na\"ive prompt  &``a photo of hazelnut''                    &1.34 &0.78 &23.46 &764.48 \\  
    \text Worst prompt  &``a {\{hazelnut\}} with {\{decantherous\}}'' &4.56 &0.79 &23.57 &782.75 \\  
    \hline
    \text Our prompt    &``a {\{hazelnut\}} with {\{cobnut\}}''   &\bf22.59  &\bf0.90 &\bf28.39 &\bf701.84 \\  
    \hline
    \hline
\end{tabular}
}
\label{tb:Evaluation of image quality}
\end{table}

\begin{figure}[t]
\centering
    \includegraphics[width=.49\linewidth]{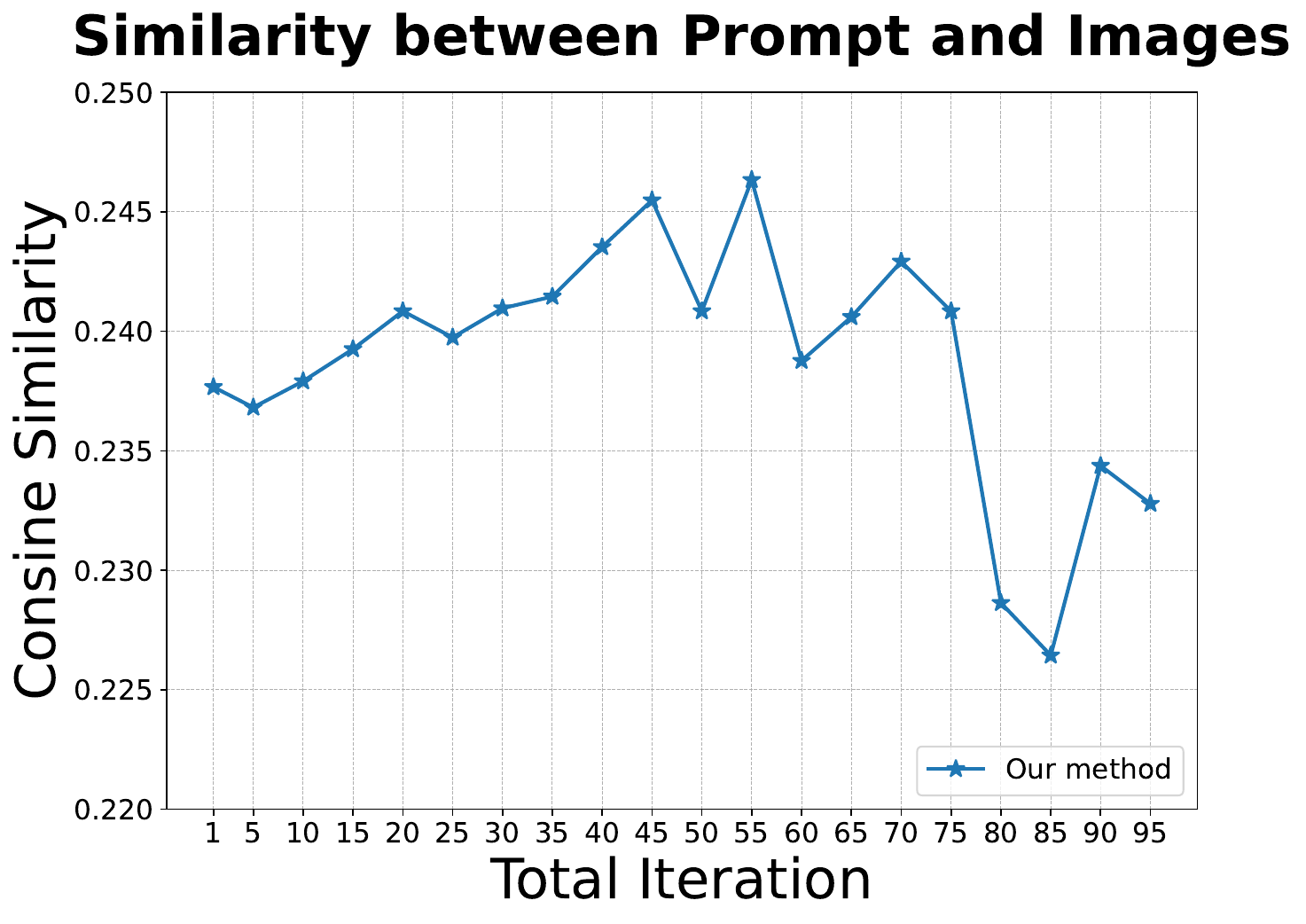}
    \includegraphics[width=.49\linewidth]{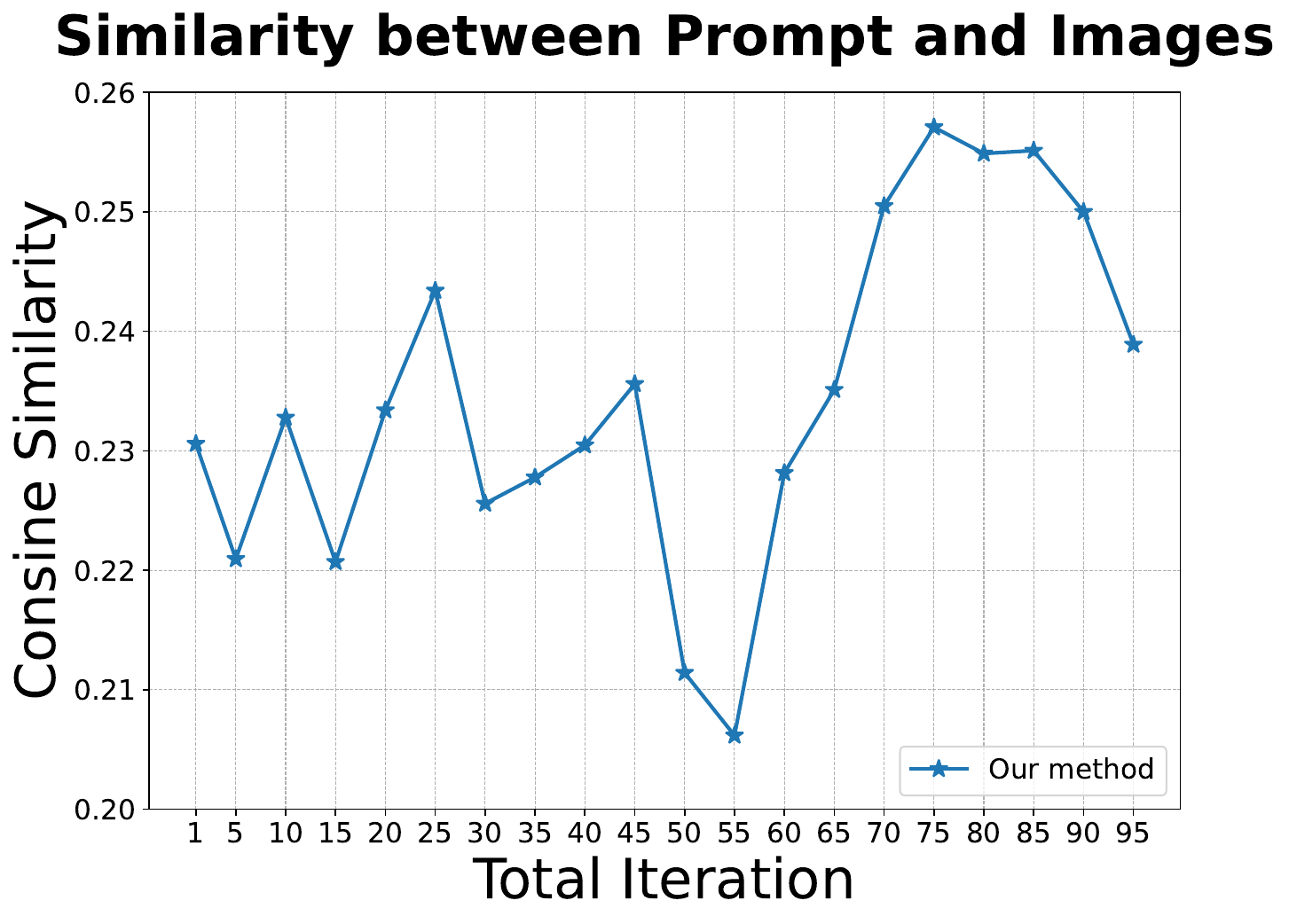}
\caption{
\textbf{The trend of similarity score between the generated image and the best prompt according to the number of iterations.} 
(Left), The trend of similarity score in hazelnut class,
(Right) The trend of similarity score in leather class in MVTec-AD dataset.}
  \label{fig:The Cosine similarity by iteration}
\end{figure}

However, failures happen in more diverse and complicated situations in the real world; thus, it is necessary to verify the robustness of our model in more complex settings.
MVTec-LOCO AD dataset is suitable for testing complex tasks since it consists of complicated classes, such as logically structured arrangement (logical type) and surface defect recognition (structural type).
Fig.~\ref{fig:Generalization test of Anomaly segmentation on MVTecAD-loco} shows the experimental results on the MVTec-LOCO AD dataset.
For the logical type, our method shows improvements of 15.5\% for the detection AUROC and 15.8\% for segmentation AUROC.
Furthermore, for the structural type, our method also shows improvements for both cases, with 10.6\% for detection AUROC and 4.3\% for segmentation AUROC.
In Fig.~\ref{fig:Generalization test of Anomaly segmentation on MVTecAD-loco}(Bottom), the qualitative results also show that our method can effectively detect and classify the changes in the placement structure of a breakfast box and the subtle differences in a juice bottle.

\subsection{Experimental Analysis}

\subsubsection{Optimizing results between our prompt and generated images}
As shown in Fig.~\ref{fig:The comparison analysis based Cosine similarity}, we visualize the results of the Keyword-to-prompt Generator successfully finding the best prompt, 
and the images generated by Variance-aware Image Generator and Text-guided Knowledge Integrator based on the best prompt.

Fig.~\ref{fig:The comparison analysis based Cosine similarity} (Top-Left) is a graph comparing the distance based on the similarity between input text and prompts based on the cosine similarity value. 
The worst prompt (blue dot) and our prompt (red dot) show a stark difference.
Fig.~\ref{fig:The comparison analysis based Cosine similarity} (Top-Right) is a graph comparing the similarity scores of input images and each prompt, and we can show that the prompt received the highest score (33\%) and the worst prompt received the lowest score (17\%).
We find that our prompts have characteristics close to the non-defective image and are evidence that they can boost the performance of text-based knowledge integration modules.
In Fig.~\ref{fig:The comparison analysis based Cosine similarity} (Bottom), we show that the qualitative image result generated by the best prompt is most similar to the original image.
In Table~\ref{tb:Evaluation of image quality}, we additionally compared the quantitative results based on the qualitative result images in Fig.~\ref{fig:The comparison analysis based Cosine similarity}.
We compared IS, SSIM, PSNR, and LPIPS scores and showed that our method had the highest diversity and high visual similarity to the original compared to other prompts.

In Fig.~\ref{fig:The Cosine similarity by iteration}, we analyze the trend of similarity between the best prompt and the generated image for each iteration.
Looking at the results, the generated image and the number of iterations are not unconditionally proportional, but the performance is optimized at a certain point and then decreases.
We prove our model is looking for optimal images while generating images with different properties.

\begin{table}[t]
\caption{\textbf{Ablation studies using Patchcore.~\cite{ref:patchcore}}
1) The performance is enhanced by the text prompt created with the text generator. 
2) Initially, performance rises as the number of generated images increases, but it eventually reaches a saturation point over time. 
3) In addition, the image generation module uses the variance method to guarantee a uniform distribution of the latent vector. This technique directly affects the quality and performance of the resulting images.}
\centering
\resizebox{\linewidth}{!}{
\begin{tabular}{l|c|c|c|c}
    \noalign{\smallskip}\noalign{\smallskip}\hline\hline
    Component                   &Image-Level (\%)&Gain (\%)&Pixel-Level (\%)&Gain (\%)   \\
    \hline
    \hline
    Baseline                    &\bf74.5($\pm$1.6)    &-    &\bf97.0($\pm$0.1)     &-      \\
    \hline
    \hline
    1) Text prompt              &                                       \\
    \hline
    w/o Text                    &77.6($\pm$0.6) &(+3.1\%) &\bf97.1($\pm$0.0) &\bf(+0.1\%)\\
    w/ Text (ours)              &\bf79.1($\pm$0.6) &\bf(+4.6\%) &96.9($\pm$0.0) &(-0.1\%)\\
    \hline
    2) N-Generated images                                                         \\
    \hline
    1-copy                      &78.6($\pm$0.4) &(+4.1\%) &97.0($\pm$0.0)  &(+0.0\%)\\
    10-copy                     &77.9($\pm$0.4) &(+3.4\%) &96.9($\pm$0.0)   &(-0.1\%)\\
    20-copy                     &78.4($\pm$0.5) &(+3.9\%) &\bf97.0($\pm$0.0)  &\bf(+0.0\%)\\
    30-copy                     &\bf79.0($\pm$0.7) &\bf(+4.5\%) &97.0($\pm$0.0)  &(+0.0\%)\\
    50-copy                     &78.5($\pm$0.3) &(+4.0\%) &96.9($\pm$0.1)  &(-0.1\%)\\
    100-copy                    &78.4($\pm$0.5) &(+3.9\%) &96.9($\pm$0.0)  &(-0.1\%)\\
    \hline
    3) Variance-aware            &                                         \\
    \hline
    w/o Variance                &74.3($\pm$0.4) &(-0.2\%) &97.0($\pm$0.0) &(+0.0\%)\\
    w/ Variance (ours)          &\bf78.9($\pm$0.5) &\bf(+4.3\%) &\bf97.3($\pm$0.0) &\bf(+0.3\%)\\
    \hline
    \hline
\end{tabular}
}
\label{tb:Ablation test}
\end{table}

\begin{figure}[t]
\centering
    \includegraphics[width=\linewidth]{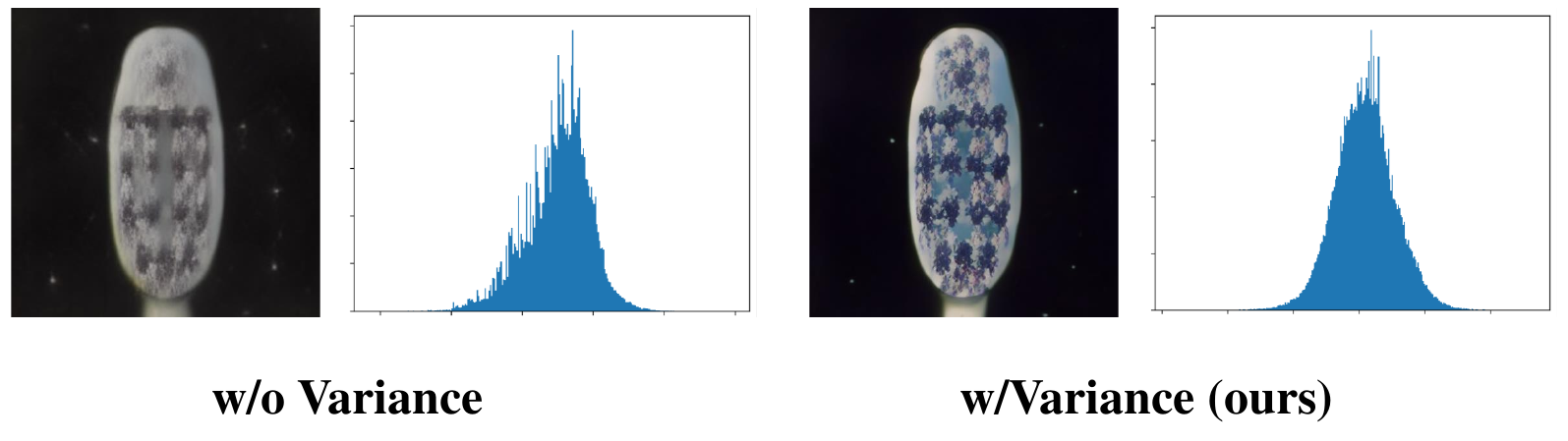}
\caption{
\textbf{Qualitative results by variance-aware method.} 
The variance-aware method affect the distribution of image latent vectors, it also affect to the quality of image generation.}
  \label{fig:Qualitative Results by variance}
\end{figure}

\subsubsection{Ablation Studies}
Table~\ref{tb:Ablation test} shows the impact of the respective component of our method.
The results with our prompt generator show 4.6\% improvement compared to the baseline.

We also analyze the effect of the quantity of the text-based generation images on the performance of the text generator.
The model's performance continuously improves as the number of text-based generation images rises and tends to reach a saturation point.
The evidence is that excessive accumulation of non-defective data can act as critical noise in data expression.
We found a performance difference of up to 1.1\% depending on the number of images.

Lastly, we compare the results of using the variance-aware method. The variance method had an impact of approximately 4.3\%.
Additionally, In Fig.~\ref{fig:Qualitative Results by variance}, we compared qualitative results for the Variance-aware method.
The Variance-aware method made latent vector from the distribution of $\mathcal{N}\{E_d, \Sigma_d\}$ value in a form close to the normal distribution, and the generation quality is also improved.

\section{Conclusion} 
To solve the issues in data shortages of the anomaly detection task in large-scale industrial manufacturing, our study proposes a new framework composed of variance-aware image generator, keyword-to-prompt generator, and text-guided knowledge integrator.
Based on diverse real-world industrial situations and data shortage scenarios, we have conducted extensive experiments using four baselines and three datasets for comparative analysis.
The experimental results show the impressive performance of our framework in all of the assumed scenarios,
especially with a single non-defective image.
We have found that the uniform distribution of the latent vector helps improve performance by preserving the characteristics of non-defective images during regeneration.
With a text-based multimodal model, our study shows our framework's potential to effectively perform anomaly detection and segmentation, even in industrial environments with insufficient data.
Furthermore, our method shows impressive performance even with a limited number of non-defective images, which enables us to avoid the problem of intertwining the defective images while effectively gathering the large-scale non-defective image set.

\vspace{1mm}
\footnotesize
\noindent\textbf{Acknowledgements:}
This work was partly supported by Institute of Information \& communications Technology Planning \& Evaluation (IITP) grant funded by the Korea government(MSIT) (2021-0-01341, Artificial Intelligence Graduate School Program(Chung-Ang University) and 2021-0-02067, Next Generation AI for Multi-purpose Video Search and 2014-3-00123, Development of High Performance Visual BigData Discovery Platform for Large-Scale Realtime Data Analysis) and a grant (22193MFDS471) from the Ministry of Food and Drug Safety in 2024.
\normalsize

{
    \small
    \bibliographystyle{ieeenat_fullname}
    \bibliography{main}
}


\clearpage

\twocolumn[{%
\renewcommand\twocolumn[1][]{#1}%
\maketitlesupplementary
\thispagestyle{empty}
\begin{center}
    \includegraphics[width=.49\linewidth]{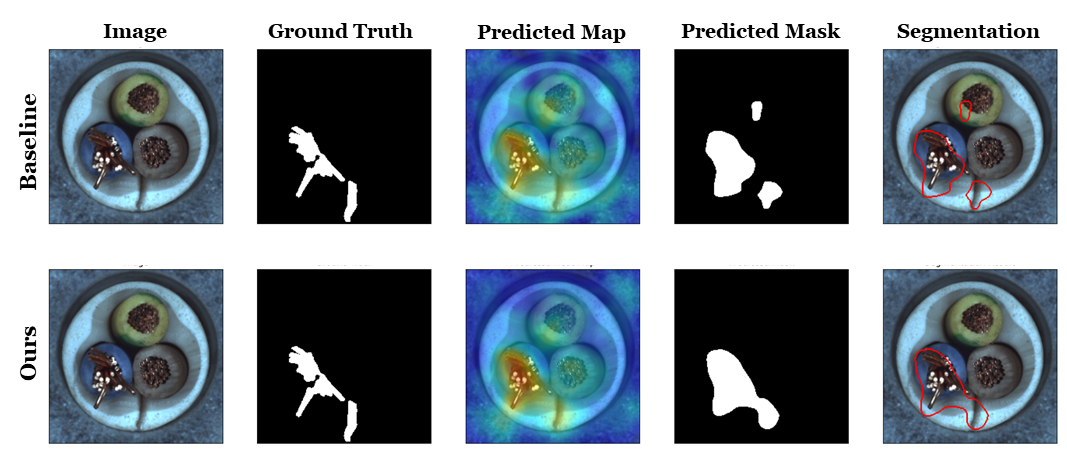}
    \includegraphics[width=.49\linewidth]{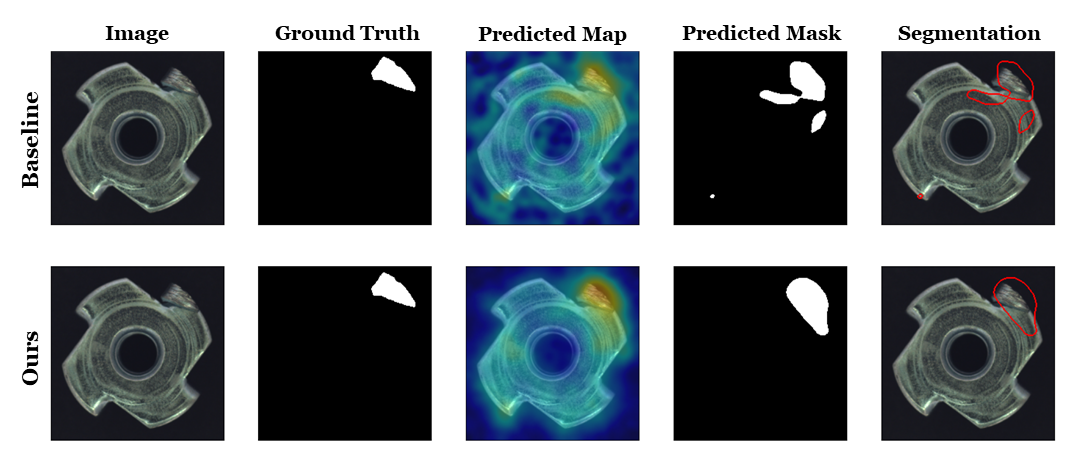}
    \includegraphics[width=.49\linewidth]{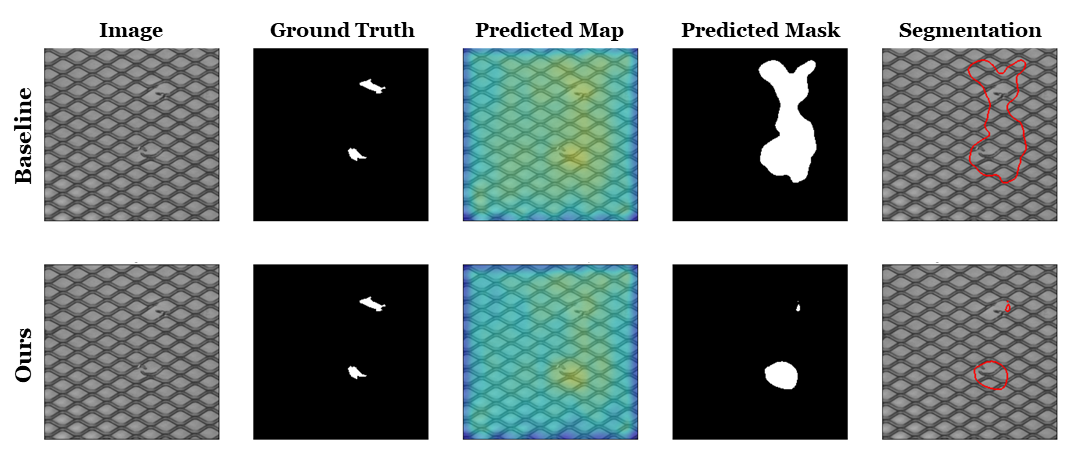}
    \includegraphics[width=.49\linewidth]{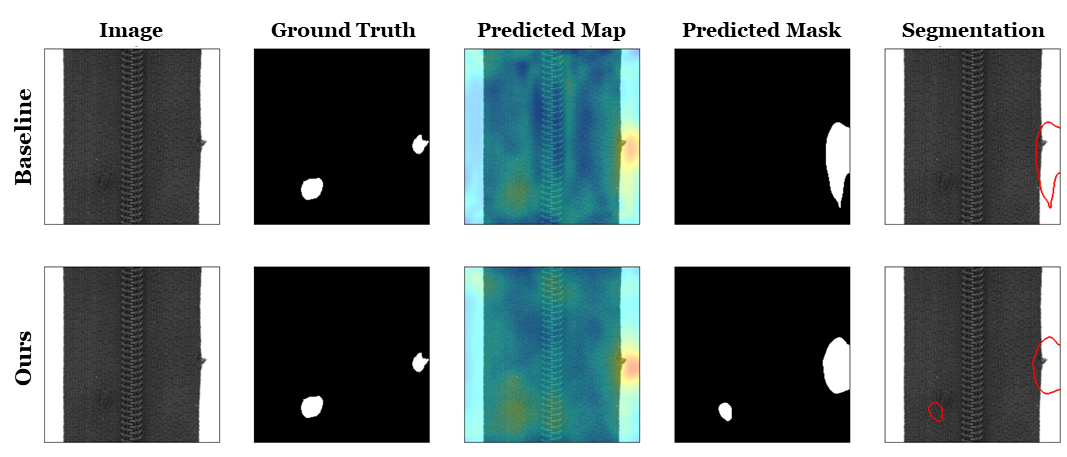}
    \includegraphics[width=.49\linewidth]{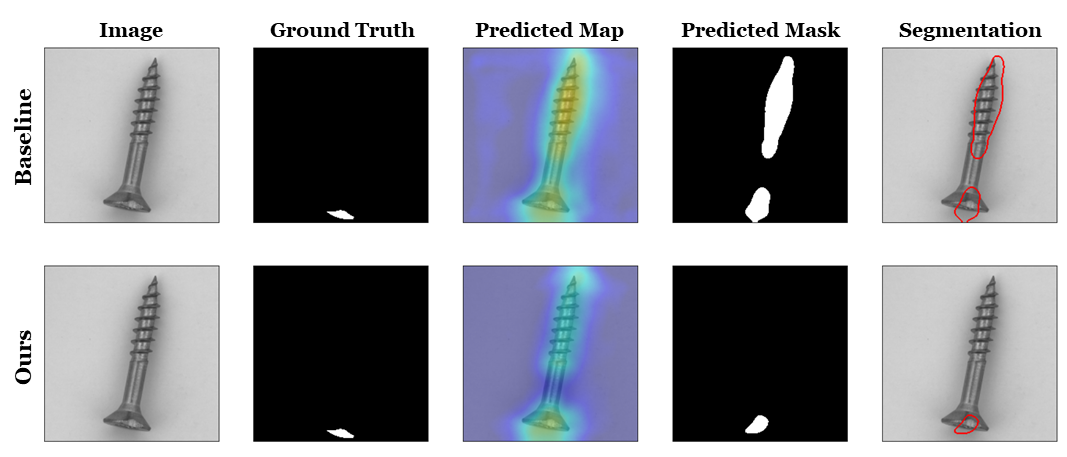}
    \includegraphics[width=.49\linewidth]{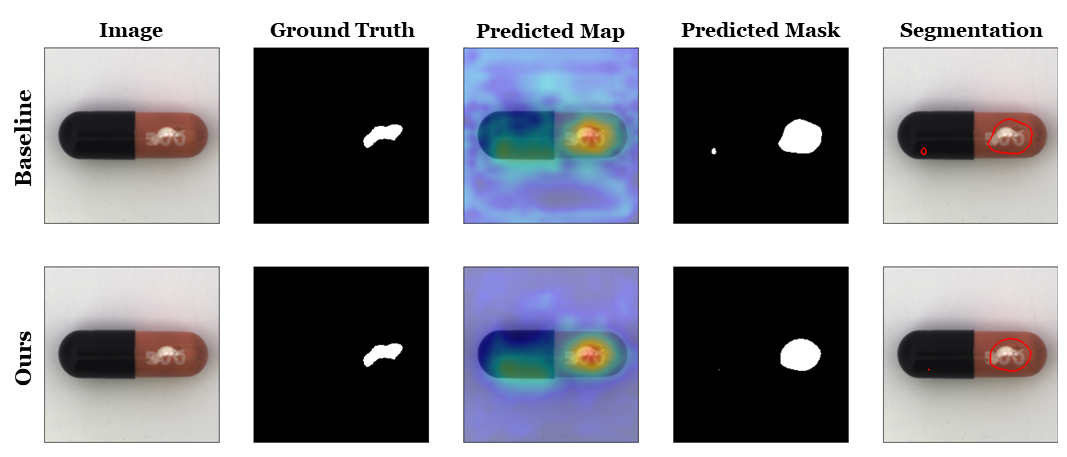}
\captionof{figure}{\textbf{Comparison of qualitative results between baseline(Patchcore) and Ours in MVTecAD dataset}. 
    Using our model, we show better detection and segmentation performance for each class compared to the baseline.} 
    \label{fig:Comparison of Qualitative results in MVTecAD dataset}
\end{center}
}]
\appendix

\section{Appendix}
Here, we present additional findings that, due to space limitations, could not be included in the main paper. 
First, detailed implementation, including evaluation metrics and baselines, benchmark datasets, were provided.
In addition, the comprehensive quantitative result (Table~\ref{tb:Performance increase rate by total percentage used original dataset}), overall classes performance results by baseline (Tables~\ref{tb:Generalization test on MVTEC-AD(Pathcore,All classes(object type)} $\sim$ ~\ref{tb:Generalization test on MVTEC-LOCO(Patchcore,All classes)}) 
and qualitative experimental results (Figs.~\ref{fig:Comparison of Qualitative results in MVTecAD dataset}, ~\ref{fig:Detection and Segmentation results[All] on MVTecAD dataset} $\sim$ ~\ref{fig:Segmentation results[All] on BTAD dataset}). 
Additional analysis results and discussion were also provided.

\subsection{Comparison of qualitative results in MVTecAD dataset.}
In Fig.~\ref{fig:Comparison of Qualitative results in MVTecAD dataset}, we can visually confirm that our method is improved over the baseline through the qualitative results of few classes.
In Table~\ref{tb:Generate Prompt by Text Generator} are few examples of prompts that were finally selected and applied.

\begin{table}[t]
\caption{\textbf{Prompt examples by Keyword-to-Prompt Generator}.}
  \centering
\centering
    \vspace{-0.5cm}
\resizebox{\linewidth}{!}{
\begin{tabular}{l|l}
    \noalign{\smallskip}\noalign{\smallskip}\hline\hline
                                         Best prompt                                     &Worst prompt                      \\
    \hline
    \hline
                                ``a {\{hazelnut}\} with {\{cobnut}\}''           &``a {\{hazelnut}\} with {\{decantherous}\}''   \\
    
                                ``a {\{metalnut}\} with {\{metallical}\}''           &``a {\{metalnut}\} with {\{predegenerate}\}''   \\
        
                                ``a {\{zipper}\} with {\{metallization}\}''           &``a {\{zipper}\} with {\{Echinops}\}''   \\
       
                                ``a {\{capsule}\} with {\{incapsulation}\}''           &``a {\{capsule}\} with {\{perceptible }\}''   \\    
     
                                ``a {\{toothbrush}\} with {\{parazoan}\}''           &``a {\{toothbrush}\} with {\{chaetopod}\}''   \\
    \hline
    \hline
\end{tabular}
}
\label{tb:Generate Prompt by Text Generator}
    \vspace{-0.5cm}
\end{table}

\subsection{Implementation Details}
The basic settings of each module are as follows:

The variance-aware image generator is initialized by the weight parameters of VQGAN~\cite{ref:vqgan} pre-trained by ImageNet.
In the Keyword-prompt generator, the number of candidate prompts $(S_1, S_2, \ldots, S_T)$ using WordNet~\cite{ref:wordnet} is 1,000; among the candidate prompts, the one closest to the input image is selected by 100 times iteration.
The text-guided knowledge integrator, the encoder, uses a pre-trained CLIP model~\cite{ref:clip} based on 'ViT-B/16'.

The experimental setup involves using the official code provided by the authors for each baseline, with results averaged over five runs. 
Each experiment was repeated 20 times with a constant learning rate of 0.05. 
VRAM usage during the experiments varied between 2,031 $\sim$ 23,277(MB), with an average runtime of 182.6(sec) per iteration. 
In all scenarios (one-shot, few-shot, and full-shot), baseline configurations were used as specified by their respective authors, with ResNet-18 serving as the default backbone network. 
And for the reverse-distillation model, a WideResNet50-2 was utilized. 
The AUROC metric was employed to evaluate both detection and segmentation performances.

\begin{figure}[t]
\centering
    \includegraphics[width=.9\linewidth]{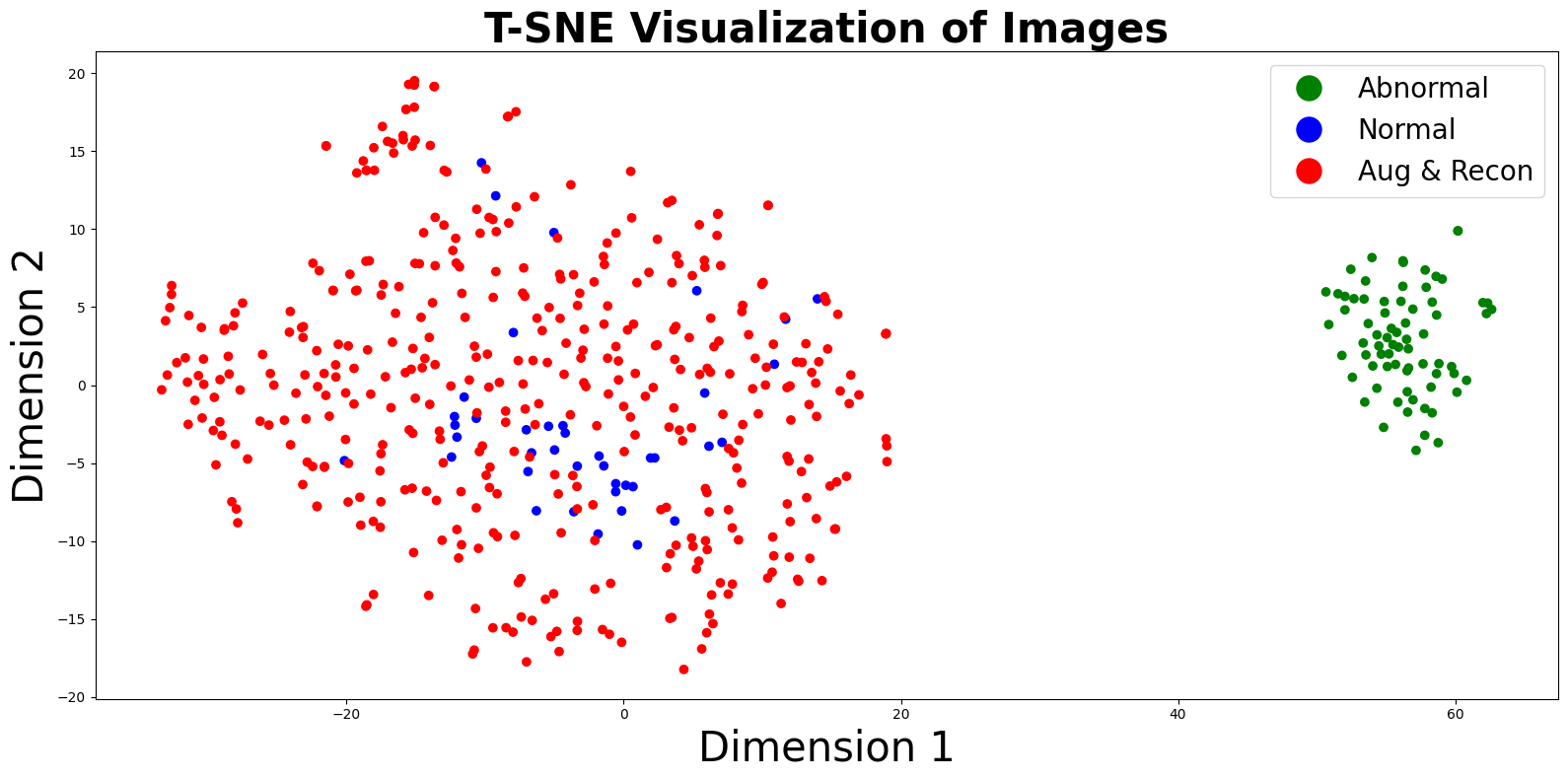}
        \vspace{-0.2cm}
  \caption{\textbf{Visual analysis on t-SNE distributions.}
  (Patchcore, Toothbrush(class)) We compare t-SNE distributions with the original images (Non-defective), generated images (Aug \& Recon), and defective images (Defective). 
  The graph shows that the generated images are generalized well with non-defective images, while effectively distinguishing the abnormal samples.}
  \label{fig:t-sne results}
        \vspace{-0.2cm}
\end{figure}

\subsubsection{Evaluation Metrics.} 
The AUROC (Area Under the Receiver Operating Characteristic curve) is a critical metric in classification models, particularly for binary classification scenarios. It measures a model's ability to differentiate between two distinct classes accurately. This metric is depicted as the area beneath the ROC curve, representing the True Positive Rate(TPR) versus the False Positive Rate(FPR) across a spectrum of threshold settings. The AUROC value, which can vary between 0 and 1, directly correlates with the model's discriminate ability; higher values indicate superior performance.

In the specific context of Anomaly Detection and Segmentation, performance evaluation extends beyond conventional metrics to include specialized tests such as the one-shot, few-shot, and full-shot tests. The one-shot test evaluates model performance using a single non-defective image and its generated counterparts, assessing its ability to detect anomalies based on minimal data. The few-shot test enhances this approach by using five non-defective images as input, providing a slightly broader base for generating and evaluating images. On the other hand, the full-shot test employs the entire dataset of non-defective images for training and image generation, offering a comprehensive evaluation of the model's capability to identify and segment anomalies.

\subsubsection{Baselines and Comparisons}
We perform a comparative analysis for four baselines.
The Patchcore~\cite{ref:patchcore} algorithm extracts features with the Coreset sampling module to use only a part of the training data. 
The Cflow~\cite{ref:cflow} uses a normalizing flow to directly predict the test image's probability by learning the invertible function mapping from images to Gaussian distribution.
The Efficient-AD~\cite{ref:efficientad} uses a student-teacher approach to detect anomalous features.
The authors of Efficient-AD train a student network to predict the extracted features of non-defective data, i.e., anomaly-free training images. 
The student's failure to predict their features is considered a detection of anomalies. 
The Reverse Distillation~\cite{ref:Reverse_Distillation} proposes a teacher encoder, student decoder, and reverse distillation paradigm model. 
Instead of receiving raw images directly, the student network takes the teacher model’s one-class embedding as input and targets to restore the teacher’s multi-scale representations.

\subsubsection{Benchmark Datasets}
We conducted a comparative analysis using three datasets to compare the models' effectiveness.
The MVTecAD (MVTec Anomaly Detection)~\cite{ref:mvtec} dataset is a benchmark dataset for anomaly detection methods focusing on industrial inspection. 
It contains over 5,000 high-resolution images divided into fifteen object and texture categories. 
Each category comprises a set of defect-free training images, a test set of images with various defects, and images without defects.
The BTAD (beanTech Anomaly Detection)~\cite{ref:btad} dataset is a real-world industrial anomaly dataset 
containing 2,830 real-world images of 3 industrial products containing body and surface defects.
The MVTec-LOCO AD~\cite{ref:mvtec_loco} dataset is intended to evaluate unsupervised anomaly localization algorithms.
It includes structural and logical anomalies, containing 3,644 images from five categories of real-world industrial inspection scenarios. The structural anomalies have scratches, dents, and contaminations in the manufactured products.
The logical anomalies violate the underlying constraints, such as an invalid location and a missing object. The dataset also includes pixel-precise ground truth data for the segmentation task.

\begin{table*}[t]
\caption{\textbf{Comprehensive results on MVTecAD(All baselines, All classes, Detection/Segmentation AUROC)}, Experimental results show improved performance in all scenarios.
All experimental results were measured with the official code provided by the author for each baseline and calculated as the average value performed five times.}
  \centering
\resizebox{.6\linewidth}{!}{
\begin{tabular}{l|c|c|c}
    \noalign{\smallskip}\noalign{\smallskip}\hline\hline
    \bf Baseline                                        &\bf One-shot  &\bf Few-shot  &\bf Full-shot\\
    \hline
    \hline
    
    Patchcore~\cite{ref:patchcore}                      &76.9\% / 90.8\%    &85.8\% / 93.2\%      &97.3\% / 97.6\%   \\
    Ours                                                &83.9\% / 92.8\%    &91.4\% / 95.1\%      &97.8\% / 97.7\%   \\
    \hline
   \textbf{Gain(+\%)}                                   &\bf+7.0\% / +2.0\%    &\bf+5.6\% / +1.9\%      &\bf+0.5\% / +0.1\%   \\
    \hline
    \hline

    Cflow~\cite{ref:cflow}                              &69.0\% / 89.0\%    &79.8\% / 92.8\%      &88.8\% / 96.1\%   \\
    Ours                                                &84.2\% / 91.6\%    &90.2\% / 94.9\%      &93.7\% / 95.9\%   \\
    \hline
    \textbf{Gain(+\%)}                                  &\bf+15.2\% / +2.6\%   &\bf+10.4\% / +2.1\%     &\bf+4.9\% / -0.2\%   \\
    \hline
    \hline

    Reverse-distillation~\cite{ref:Reverse_Distillation}&48.4\% / 29.4\%    &50.6\% / 32.7\%      &79.2\% / 97.0\%   \\
    Ours                                                &83.1\% / 93.3\%    &84.2\% / 92.1\%      &93.3\% / 97.1\%   \\
    \hline
    \textbf{Gain(+\%)}                                  &\bf+34.7\% / +63.9\%  &\bf+33.7\% / +59.4\%    &\bf+14.1\% / +0.1\%   \\
    \hline
    \hline

    Efficient-AD~\cite{ref:efficientad}                 &65.6\% / 78.1\%    &71.3\% / 83.3\%      &92.8\% / 93.6\%   \\
    Ours                                                &71.3\% / 83.7\%    &78.1\% / 85.9\%      &95.5\% / 94.3\%   \\
    \hline
    \textbf{Gain(+\%)}                                  &\bf+5.8\% / +5.6\%    &\bf+6.8\% / +2.6\%      &\bf+2.7\% / +0.7\%    \\

    \hline
    \hline
    \bf Average                                         &\bf+15.7\% / +18.6\%  &\bf+14.1\% / +16.5\% &\bf+5.6\% / +0.2\%    \\
     
    \hline
    \hline
\end{tabular}
}
\label{tb:Performance increase rate by total percentage used original dataset}
\end{table*}

\begin{figure*}[t]
\centering
    \includegraphics[width=\linewidth]{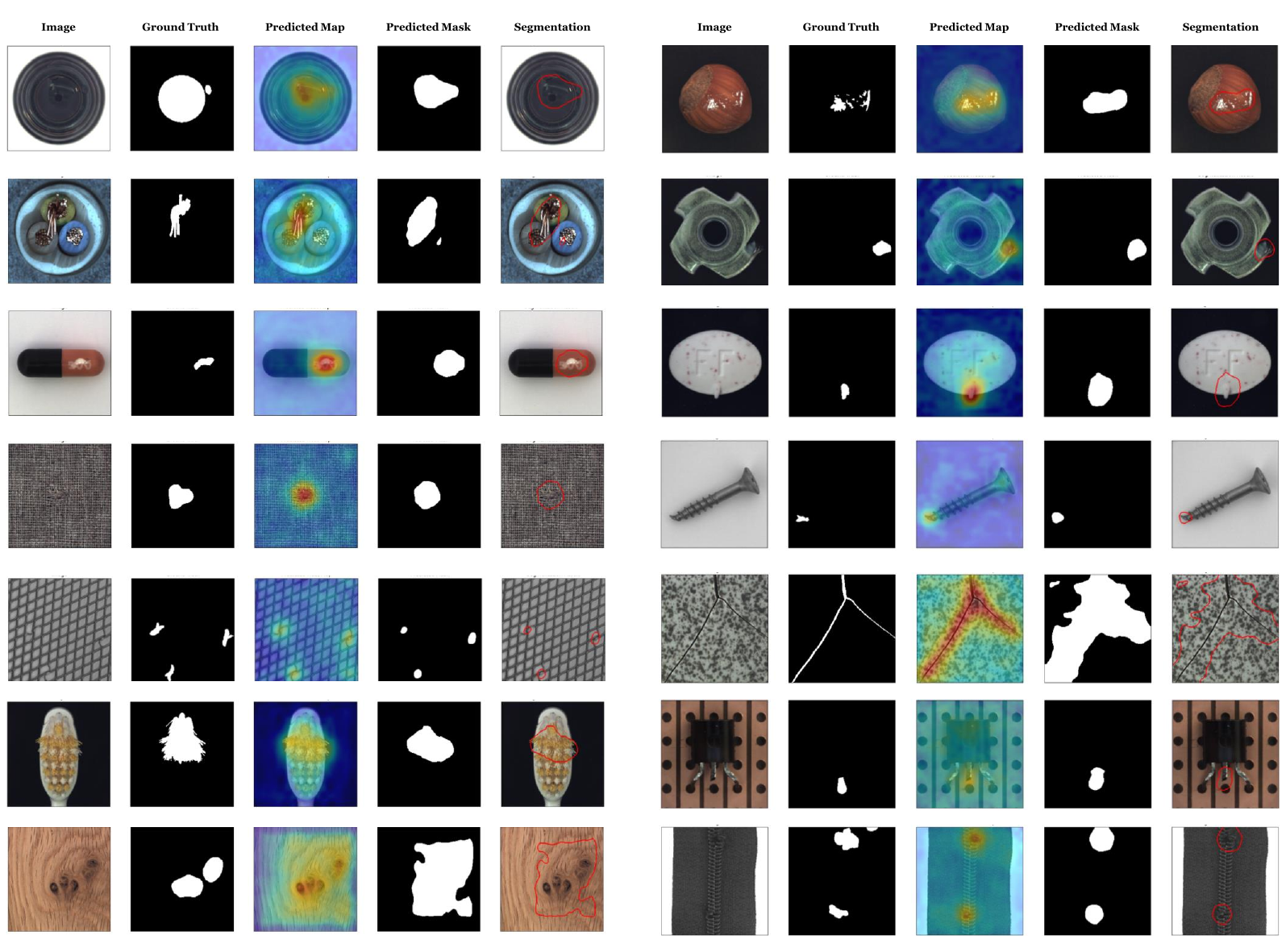}
  \caption{\textbf{Qualitative results[All classes] on MVTecAD dataset}. 
For each class in the scenario test, we show the qualitative results of defective images using our model. 
}
  \label{fig:Detection and Segmentation results[All] on MVTecAD dataset}
\end{figure*}

\begin{table*}[ht!]
\caption{\textbf{Generalization test on MVTecAD(Patchcore, Object type classes, Detection AUROC)},
Generalization test on MVTec-AD dataset~\cite{ref:mvtec}, and the values of object type among all classes in the patchcore~\cite{ref:patchcore} baseline and their average values are expressed.
All experimental results were measured using the official code provided by the author for each baseline and calculated as the average value performed five times.
}
  \centering
\centering
    \vspace{-0.3cm}
\resizebox{\linewidth}{!}{
\begin{tabular}{l|cccccccccc|c}
    \noalign{\smallskip}\noalign{\smallskip}\hline\hline
                    \bf{}& Bottle& Cable& Capsule& Hazelnut& Metal-nut& Pill& Screw& Toothbrush& Transistor& Zipper& Avg\\
    \hline
    One-shot &99.5($\pm$0.2) &70.8($\pm$1.1) &55.5($\pm$1.6) &88.3($\pm$0.2) &70.3($\pm$0.7) &69.2($\pm$2.4) &48.1($\pm$1.4) &74.7($\pm$0.7) &65.3($\pm$3.0) &78.3($\pm$1.2) &72.0 \\
    
    Ours  &100.0($\pm$0.0) &89.9($\pm$0.3) &63.6($\pm$0.2) &91.7($\pm$0.1) &88.8($\pm$0.2) &72.0($\pm$0.4) &53.8($\pm$0.1) &78.7($\pm$0.4) &71.8($\pm$0.7) &91.5($\pm$0.2) &\bf80.2 \\
    \hline    
   \textbf{Gain(+\%)}  &\bf+0.5\% &\bf+19.1\% &\bf+8.1\% &\bf+3.4\% &\bf+18.5\% &\bf+2.8\% &\bf+5.7\% &\bf+4.0\% &\bf+6.5\% &\bf+13.2\% &\bf+8.2\%\\
    \hline    
    \hline    

    Few-shot &99.3($\pm$0.1) &87.7($\pm$0.9) &77.7($\pm$1.8) &96.0($\pm$1.1) &95.3($\pm$0.4) &84.3($\pm$2.7) &50.3($\pm$0.3) &63.3($\pm$0.7) &93.5($\pm$0.3) &91.9($\pm$0.3) &83.9 \\
    
    Ours   &100.0($\pm$0.0) &95.6($\pm$0.8) &93.0($\pm$0.4) &99.3($\pm$0.1) &99.1($\pm$0.1) &88.9($\pm$4.3) &66.6($\pm$0.3) &69.5($\pm$0.2) &98.9($\pm$0.2) &92.4($\pm$0.1) &\bf90.3\\
    \hline

   \textbf{Gain(+\%)}  &\bf+0.7\% &\bf+7.9\% &\bf+15.3\% &\bf+3.3\% &\bf+3.8\% &\bf+4.6\% &\bf+16.3\% &\bf+6.2\% &\bf+5.4\% &\bf+0.5\% &\bf+6.4\%\\
    \hline
    \hline    

    Full-shot &100.0($\pm$0.0) &98.6($\pm$0.1) &96.5($\pm$0.6) &100.0($\pm$0.0) &99.1($\pm$0.3) &91.6($\pm$1.5) &94.3($\pm$0.6) &93.1($\pm$1.1) &99.6($\pm$0.2) &95.3($\pm$0.1) &96.8 \\
    
    Ours  &100.0($\pm$0.0) &98.8($\pm$0.1) &97.2($\pm$0.3) &100.0($\pm$0.0) &100.0($\pm$0.0) &92.5($\pm$0.2) &96.1($\pm$0.5) &95.7($\pm$0.9) &100.0($\pm$0.0) &95.7($\pm$0.1) &\bf{97.6}\\
    \hline

   \textbf{Gain(+\%)}  &\bf+0.0\% &\bf+0.2\% &\bf+0.7\% &\bf+0.0\% &\bf+0.9\% &\bf+0.9\% &\bf+1.8\% &\bf+2.6\% &\bf+0.4\% &\bf+0.4\% &\bf+0.8\%\\  
   
    \hline
    \hline  
\end{tabular}
}
\label{tb:Generalization test on MVTEC-AD(Pathcore,All classes(object type)}
\end{table*}

\begin{table*}[ht!]
\caption{\textbf{Generalization test on MVTecAD(Patchcore, Texture type classes, Detection AUROC)},
Generalization test on MVTec-AD dataset~\cite{ref:mvtec}, and the values of texture type among all classes in the patchcore~\cite{ref:patchcore} baseline and their average values are expressed.
All experimental results were measured using the official code provided by the author for each baseline and calculated as the average value performed five times.
}
  \centering
\centering
    \vspace{-0.3cm}
\resizebox{.6\linewidth}{!}{
\begin{tabular}{l|ccccc|c}
    \noalign{\smallskip}\noalign{\smallskip}\hline\hline
                    \bf{} &Carpet &Grid &Leather &Tile &Wood &Avg \\
    \hline
    One-shot            &89.4($\pm$2.2) &47.2($\pm$0.2) &99.9($\pm$0.1) &99.3($\pm$0.2) &97.6($\pm$0.3) &86.7 \\
    Ours                &93.6($\pm$0.5) &64.9($\pm$0.2) &100.0($\pm$0.0) &99.3($\pm$0.1) &98.9($\pm$0.1) &\bf91.3 \\
    \hline 
    \textbf{Gain(+\%)}  &\bf+4.2\% &\bf+17.7\% &\bf+0.1\% &\bf+0.0\% &\bf+1.3\% &\bf+4.6\%\\
    \hline 
    \hline 
    Few-shot            &93.4($\pm$0.7) &56.4($\pm$2.9) &100.0($\pm$0.0) &99.4($\pm$0.2) &98.7($\pm$0.1) &89.6 \\
    Ours                &96.4($\pm$0.3) &72.3($\pm$0.8) &100.0($\pm$0.0) &99.8($\pm$0.0) &99.5($\pm$0.0) &\bf93.6\\
    \hline
    \textbf{Gain(+\%)}  &\bf+3.0\% &\bf+15.9\% &\bf+0.0\% &\bf+0.4\% &\bf+0.8\% &\bf+4.0\%\\
    \hline 
    \hline
    Full-shot           &97.0($\pm$0.5) &94.7($\pm$0.4) &100.0($\pm$0.0) &99.7($\pm$0.1) &99.7($\pm$0.1) &98.2 \\
    Ours                &96.7($\pm$0.2) &94.3($\pm$0.4) &100.0($\pm$0.0) &99.9($\pm$0.0) &99.7($\pm$0.0) &\bf98.1\\
    \hline
    \textbf{Gain(+\%)}  &\bf-0.3\% &\bf-0.4\% &\bf+0.0\% &\bf+0.2\% &\bf+0.0\% &\bf-0.1\%\\
    \hline  
    \hline  
\end{tabular}
}
\label{tb:Generalization test on MVTEC-AD(Pathcore,All classes(texture type)}
\end{table*}

\begin{table*}[ht!]
\caption{\textbf{Generalization test on MVTecAD(Cflow, Object type classes, Detection AUROC)},
Generalizationtion test on MVTec-AD dataset~\cite{ref:mvtec}, and the values of Object type among all classes in the cflow~\cite{ref:cflow} baseline and their average values are expressed.
All experimental results were measured using the official code provided by the author for each baseline and calculated as the average value performed five times.
}
  \centering
\centering
    \vspace{-0.3cm}
\resizebox{\linewidth}{!}{
\begin{tabular}{l|cccccccccc|c}
    \noalign{\smallskip}\noalign{\smallskip}\hline\hline
                    \bf{}& Bottle& Cable& Capsule& Hazelnut& Metal-nut& Pill& Screw& Toothbrush& Transistor& Zipper& Avg \\
    \hline
    One-shot &93.7($\pm$1.1) &57.8($\pm$6.9) &71.5($\pm$0.7) &90.5($\pm$0.7) &60.1($\pm$1.4) &68.6($\pm$3.2) &54.6($\pm$1.1) &68.3($\pm$2.3) &61.8($\pm$3.0) &52.3($\pm$3.4) &67.9 \\
    Ours  &99.1($\pm$0.9) &77.7($\pm$1.9) &73.6($\pm$3.3) &96.4($\pm$1.3) &79.6($\pm$3.0) &61.0($\pm$6.5) &79.6($\pm$6.4) &71.1($\pm$1.5) &73.7($\pm$5.1) &92.1($\pm$1.5) &\bf80.4\\
    \hline    
   \textbf{Gain(+\%)}  &\bf+5.4\% &\bf+19.9\% &\bf+2.1\% &\bf+5.9\% &\bf+19.5\% &\bf-7.6\% &\bf+25.0\% &\bf+2.8\% &\bf+11.9\% &\bf+39.8\% &\bf+12.5\%\\
    \hline
    \hline
    Few-shot &96.2($\pm$0.3) &79.9($\pm$1.3) &68.7($\pm$1.4) &96.4($\pm$0.6) &75.8($\pm$2.9) &73.6($\pm$4.1) &50.4($\pm$3.6) &82.3($\pm$3.9) &68.2($\pm$3.1) &73.3($\pm$3.9) &76.5 \\
    Ours   &99.9($\pm$0.1) &92.5($\pm$0.9) &77.6($\pm$5.1) &99.5($\pm$0.3) &92.1($\pm$2.1) &88.0($\pm$3.1) &61.7($\pm$7.5) &93.9($\pm$1.6) &86.5($\pm$1.6) &92.5($\pm$0.7) &\bf88.4\\
    \hline
    \hline    
   \textbf{Gain(+\%)}  &\bf+3.7\% &\bf+12.6\% &\bf+8.9\% &\bf+3.1\% &\bf+16.3\% &\bf+14.4\% &\bf+11.3\% &\bf+11.6\% &\bf+18.3\% &\bf+19.2\% &\bf+11.9\%\\
    \hline
    \hline
    
    Full-shot &100.0($\pm$0.0) &89.4($\pm$1.9) &80.6($\pm$1.9) &99.1($\pm$0.6) &97.9($\pm$0.6) &84.2($\pm$5.8) &55.6($\pm$4.3) &87.7($\pm$1.3) &84.0($\pm$0.9) &91.9($\pm$0.6) &87.0 \\
    Ours  &100.0($\pm$0.0) &95.7($\pm$0.9) &91.6($\pm$1.6) &99.9($\pm$0.1) &99.2($\pm$0.4) &91.2($\pm$1.2) &74.9($\pm$4.0) &91.6($\pm$1.3) &91.2($\pm$4.8) &95.0($\pm$0.9) &\bf{93.0}\\
    \hline    
   \textbf{Gain(+\%)}  &\bf+0.0\% &\bf+6.3\% &\bf+11.0\% &\bf+0.8\% &\bf+1.3\% &\bf+7.0\% &\bf+19.3\% &\bf+3.9\% &\bf+7.2\% &\bf+3.1\% &\bf+6.0\%\\
    \hline
    \hline  
\end{tabular}
}
\label{tb:Generalization test on MVTEC-AD(Cflow,All classes(object type)}
\end{table*}

\begin{table*}[ht!]
\caption{\textbf{Generalization test on MVTEC-AD(Cflow, Texture type classes, Detection AUROC)},
Generalizationtion test on MVTec-AD dataset~\cite{ref:mvtec}, and the values of texture type among all classes in the cflow~\cite{ref:cflow} baseline and their average values are expressed.
All experimental results were measured using the official code provided by the author for each baseline and calculated as the average value performed five times.
}
  \centering
\centering
    \vspace{-0.3cm}
\resizebox{.6\linewidth}{!}{
\begin{tabular}{l|ccccc|c}
    \noalign{\smallskip}\noalign{\smallskip}\hline\hline
                    \bf{} &Carpet &Grid &Leather &Tile &Wood &Avg \\
    \hline
    One-shot &88.2($\pm$4.3) &33.9($\pm$1.8) &92.7($\pm$0.7) &94.0($\pm$0.6) &96.2($\pm$2.1) &81.0 \\
    Ours  &96.8($\pm$1.4) &53.8($\pm$11.4) &94.6($\pm$2.3) &96.5($\pm$0.7) &99.0($\pm$0.2) &\bf88.1\\
    \hline 
    \textbf{Gain(+\%)}  &\bf+8.6\% &\bf+19.9\% &\bf+1.9\% &\bf+2.5\% &\bf+2.8\% &\bf+7.1\%\\
    \hline 
    \hline 
    Few-shot &91.5($\pm$1.6) &55.5($\pm$0.9) &95.5($\pm$1.3) &96.0($\pm$0.6) &93.7($\pm$3.0) &86.4 \\\
    Ours   &95.8($\pm$0.7) &77.7($\pm$2.3) &99.1($\pm$0.5) &98.3($\pm$0.6) &97.9($\pm$0.5) &\bf93.8\\
    \hline 
    \textbf{Gain(+\%)}  &\bf+4.3\% &\bf+22.2\% &\bf+3.6\% &\bf+2.3\% &\bf+4.2\% &\bf+7.4\%\\
    \hline 
    \hline 
    Full-shot &94.1($\pm$0.4) &76.4($\pm$1.9) &98.0($\pm$0.4) &96.0($\pm$2.8) &96.8($\pm$2.0) &92.3 \\
    Ours  &95.2($\pm$0.8) &84.1($\pm$4.2) &98.6($\pm$0.3) &98.2($\pm$0.7) &98.8($\pm$0.4) &\bf{95.0}\\
    \hline
    \textbf{Gain(+\%)}  &\bf+1.1\% &\bf+7.7\% &\bf+0.6\% &\bf+2.2\% &\bf+2.0\% &\bf+2.7\%\\
    \hline 
    \hline
\end{tabular}
}
\label{tb:Generalization test on MVTEC-AD(Cflow,All classes(texture type)}
\end{table*}

\begin{table*}[ht!]
\caption{\textbf{Generalization test on MVTecAD(Reverse-distillation, Object type classes, Detection AUROC)},
Generalization test on MVTec-AD dataset~\cite{ref:mvtec}, and the values of object type among all classes in the reverse-distillation~\cite{ref:Reverse_Distillation} baseline and their average values are expressed. 
All experimental results were measured using the official code provided by the author for each baseline and calculated as the average value performed five times.
}
  \centering
\centering
    \vspace{-0.3cm}
\resizebox{\linewidth}{!}{
\begin{tabular}{l|cccccccccc|c}
    \noalign{\smallskip}\noalign{\smallskip}\hline\hline
                    \bf{}& Bottle& Cable& Capsule& Hazelnut& Metal-nut& Pill& Screw& Toothbrush& Transistor& Zipper& Avg \\
    \hline
    One-shot &44.1($\pm$0.5) &47.9($\pm$0.1) &58.8($\pm$0.0) &37.2($\pm$0.2) &45.4($\pm$12.7) &57.1($\pm$0.3) &53.7($\pm$1.8) &39.0($\pm$0.2) &55.1($\pm$7.1) &41.9($\pm$1.0) &48.0 \\
    Ours     &98.0($\pm$1.8) &65.0($\pm$1.0) &67.4($\pm$2.2) &99.9($\pm$0.1) &62.7($\pm$2.1) &81.2($\pm$0.9) &54.0($\pm$1.2) &92.0($\pm$0.6) &70.4($\pm$0.8) &78.0($\pm$1.1) &\bf76.9 \\
    \hline    
   \textbf{Gain(+\%)}  &\bf+53.9\% &\bf+17.1\% &\bf+8.6\% &\bf+62.7\% &\bf+17.3\% &\bf+24.1\% &\bf+0.3\% &\bf+53.0\% &\bf+15.3\% &\bf+36.1\% &\bf+28.9\%\\
    \hline
    \hline     
    Few-shot &55.2($\pm$0.0) &54.1($\pm$0.0) &53.2($\pm$0.0) &35.7($\pm$0.0) &40.7($\pm$0.0) &53.1($\pm$0.0) &54.7($\pm$0.0) &44.7($\pm$0.0) &60.5($\pm$14.6) &34.6($\pm$0.0) &48.6 \\
    Ours     &98.6($\pm$1.5) &84.6($\pm$3.3) &74.7($\pm$1.0) &100.0($\pm$0.0) &79.9($\pm$11.8) &79.8($\pm$0.7) &63.6($\pm$0.9) &94.2($\pm$0.4) &84.0($\pm$2.2) &73.7($\pm$0.7) &\bf83.3\\
    \hline    
   \textbf{Gain(+\%)}  &\bf+43.4\% &\bf+30.5\% &\bf+21.5\% &\bf+64.3\% &\bf+39.2\% &\bf+26.7\% &\bf+8.9\% &\bf+49.5\% &\bf+23.5\% &\bf+39.1\%  &\bf+34.7\% \\
    \hline
    \hline  
    Full-shot &95.3($\pm$3.8) &95.0($\pm$0.7) &88.5($\pm$4.0) &100.0($\pm$0.0) &61.0($\pm$2.4) &59.4($\pm$0.5) &93.4($\pm$1.7) &75.9($\pm$10.8) &80.1($\pm$7.9) &89.8($\pm$0.9) &83.8 \\
    Ours      &98.7($\pm$1.8) &93.9($\pm$1.7) &94.4($\pm$0.7) &100.0($\pm$0.0) &75.1($\pm$17.8) &95.4($\pm$2.5) &95.7($\pm$1.3) &96.6($\pm$0.6) &97.1($\pm$0.6) &92.1($\pm$1.7) &\bf{93.9} \\
    \hline    
   \textbf{Gain(+\%)}  &\bf+3.4\% &\bf-1.1\% &\bf+5.9\% &\bf+0.0\% &\bf+14.1\% &\bf+36.0\% &\bf+2.3\% &\bf+20.7\% &\bf+17.0\% &\bf+2.3\% &\bf+10.1\%\\
    \hline
    \hline  
\end{tabular}
}
\label{tb:Generalization test on MVTEC-AD(Reverse-distillation,All classes(object type)}
\end{table*}

\begin{table*}[ht!]
\caption{\textbf{Generalization test on MVTecAD(Reverse-distillation, Texture type classes, Detection AUROC)},
Generalization test on MVTec-AD dataset~\cite{ref:mvtec}, and the values of texture type among all classes in the reverse-distillation~\cite{ref:Reverse_Distillation} baseline and their average values are expressed.
All experimental results were measured using the official code provided by the author for each baseline and calculated as the average value performed five times.
}
  \centering
\centering
    \vspace{-0.3cm}
\resizebox{.6\linewidth}{!}{
\begin{tabular}{l|ccccc|c}
    \noalign{\smallskip}\noalign{\smallskip}\hline\hline
                    \bf{} &Carpet &Grid &Leather &Tile &Wood &Avg\\
    \hline
    One-shot &63.0($\pm$32.9) &56.9($\pm$0.1) &26.7($\pm$0.2) &45.7($\pm$0.5) &54.3($\pm$0.5) &49.3 \\
    Ours     &99.4($\pm$0.0) &79.7($\pm$6.7) &100.0($\pm$0.0) &99.4($\pm$0.1) &99.6($\pm$0.0) &\bf95.6 \\
    \hline
    \textbf{Gain(+\%)}  &\bf+36.4\% &\bf+22.8\% &\bf+73.3\% &\bf+53.7\% &\bf+45.3\% &\bf+46.3\%\\
    \hline 
    \hline 
    Few-shot &35.3($\pm$0.0) &52.1($\pm$0.0) &42.4($\pm$1.0) &23.4($\pm$0.0) &53.3($\pm$0.0) &41.3 \\
    Ours     &99.4($\pm$0.1) &85.1($\pm$1.7) &99.9($\pm$0.0) &98.0($\pm$0.3) &99.8($\pm$0.1) &\bf96.4\\
    \hline
    \textbf{Gain(+\%)}  &\bf+64.1\% &\bf+33.0\% &\bf+57.5\% &\bf+74.6\% &\bf+46.5\% &\bf+55.1\%\\
    \hline 
    \hline
    Full-shot &99.4($\pm$0.0) &96.6($\pm$1.6) &39.6($\pm$0.5) &78.0($\pm$23.0) &99.7($\pm$0.0) &82.7 \\
    Ours      &99.5($\pm$0.1) &99.5($\pm$0.2) &100.0($\pm$0.0) &99.6($\pm$0.1) &99.7($\pm$0.0) &\bf{99.7}\\
    \hline
    \textbf{Gain(+\%)}  &\bf+0.1\% &\bf+2.9\% &\bf+60.4\% &\bf+21.6\% &\bf+0.0\% &\bf+17.0\%\\
    \hline 
    \hline
\end{tabular}
}
\label{tb:Generalization test on MVTEC-AD(Reverse-distillation,All classes(texture type)}
\end{table*}

\begin{table*}[ht!]
\caption{\textbf{Generalization test on MVTecAD(Efficient-AD, Object type classes, Detection AUROC)},
Generalization test on  MVTec-AD dataset~\cite{ref:mvtec}, and the values of object type among all classes in the Efficient-AD~\cite{ref:efficientad} baseline and their average values are expressed.
All experimental results were measured using the official code provided by the author for each baseline and calculated as the average value performed five times.
}
  \centering
\centering
    \vspace{-0.3cm}
\resizebox{\linewidth}{!}{
\begin{tabular}{l|cccccccccc|c}
    \noalign{\smallskip}\noalign{\smallskip}\hline\hline
                    \bf{}& Bottle& Cable& Capsule& Hazelnut& Metal-nut& Pill& Screw& Toothbrush& Transistor& Zipper& Avg \\
    \hline
    One-shot &84.0($\pm$0.0) &50.2($\pm$0.1) &39.6($\pm$0.0) &72.8($\pm$0.0) &37.4($\pm$0.0) &69.6($\pm$0.1) &66.4($\pm$0.0) &42.8($\pm$0.1) &34.9($\pm$0.0) &56.3($\pm$0.0) &55.4 \\
    Ours     &95.1($\pm$0.7) &61.9($\pm$0.7) &51.7($\pm$0.6) &75.9($\pm$0.8) &62.1($\pm$0.6) &73.2($\pm$2.3) &74.3($\pm$1.1) &63.4($\pm$1.7) &58.5($\pm$1.3) &52.8($\pm$0.2) &\bf66.9 \\
    \hline 
    \textbf{Gain(+\%)}  &\bf+11.1\% &\bf+11.7\% &\bf+12.1\% &\bf+3.1\% &\bf+24.7\% &\bf+3.6\% &\bf+7.9\% &\bf+20.6\% &\bf+23.6\% &\bf-3.5\% &\bf+11.5\% \\
    \hline 
    \hline 
    Few-shot &96.2($\pm$0.1) &60.5($\pm$0.2) &45.2($\pm$0.3) &69.5($\pm$0.2) &52.6($\pm$0.4) &70.4($\pm$0.7) &63.8($\pm$0.1) &51.2($\pm$0.4) &43.3($\pm$0.1) &54.9($\pm$0.1) &60.8 \\
    Ours     &98.0($\pm$0.1) &75.9($\pm$0.4) &55.2($\pm$0.2) &87.3($\pm$0.7) &67.9($\pm$1.9) &75.1($\pm$1.0) &75.4($\pm$3.7) &69.2($\pm$0.9) &60.8($\pm$1.4) &57.6($\pm$1.5) &\bf72.2 \\
    \hline
    \textbf{Gain(+\%)}  &\bf+1.8\% &\bf+15.4\% &\bf+10.0\% &\bf+17.8\% &\bf+15.3\% &\bf+4.7\% &\bf+11.6\% &\bf+18.0\% &\bf+17.5\% &\bf+2.7\% &\bf+11.4\% \\
    \hline 
    \hline 
    Full-shot &100.0($\pm$0.0) &91.9($\pm$0.2) &76.2($\pm$1.2) &89.1($\pm$0.8) &96.7($\pm$0.1) &95.5($\pm$1.0) &90.3($\pm$0.6) &94.8($\pm$0.5) &82.4($\pm$2.1) &94.6($\pm$0.4) &91.2 \\
    Ours      &100.0($\pm$0.0) &94.7($\pm$0.3) &80.7($\pm$2.6) &94.6($\pm$0.8) &97.0($\pm$0.1) &97.5($\pm$0.6) &95.1($\pm$0.5) &89.3($\pm$0.8) &82.5($\pm$3.6) &94.4($\pm$0.8) &\bf{92.6} \\
    \hline
    \textbf{Gain(+\%)}  &\bf+0.0\% &\bf+2.8\% &\bf+4.5\% &\bf+5.5\% &\bf+0.3\% &\bf+2.0\% &\bf+4.8\% &\bf-5.5\% &\bf+0.1\% &\bf-0.2\% &\bf+1.4\% \\
    \hline 
    \hline 
\end{tabular}
}
\label{tb:Generalization test on MVTEC-AD(Efficient-AD,All classes(object type)}
\end{table*}

\begin{table*}[ht!]
\caption{\textbf{Generalization test on MVTecAD(Efficient-AD, Texture type classes, Detection AUROC)},
Generalization test on MVTec-AD dataset~\cite{ref:mvtec}, and the values of texture type among all classes in the Efficient-AD~\cite{ref:efficientad} baseline and their average values are expressed. 
All experimental results were measured using the official code provided by the author for each baseline and calculated as the average value performed five times.
}
  \centering
\centering
    \vspace{-0.3cm}
\resizebox{.6\linewidth}{!}{
\begin{tabular}{l|ccccc|c}
    \noalign{\smallskip}\noalign{\smallskip}\hline\hline
                    \bf{} &Carpet &Grid &Leather &Tile &Wood &Avg \\
    \hline
    One-shot &99.4($\pm$0.0) &83.2($\pm$0.0) &65.8($\pm$0.1) &95.2($\pm$0.0) &85.7($\pm$0.0) &85.9 \\
    Ours     &99.5($\pm$0.1) &74.3($\pm$4.8) &54.8($\pm$8.6) &97.4($\pm$0.1) &81.4($\pm$0.9) &\bf81.5 \\
    \hline 
    \textbf{Gain(+\%)}  &\bf+0.1\% &\bf-8.9\% &\bf-11.0\% &\bf+2.2\% &\bf-4.3\% &\bf-4.4\%\\
    \hline 
    \hline 
    Few-shot &98.4($\pm$0.1) &94.2($\pm$0.1) &91.4($\pm$4.7) &94.3($\pm$0.2) &82.9($\pm$0.4) &92.2 \\
    Ours     &97.8($\pm$0.3) &97.9($\pm$0.4) &66.3($\pm$1.2) &96.2($\pm$0.9) &90.7($\pm$0.5) &\bf89.8 \\
    \hline
    \textbf{Gain(+\%)}  &\bf-0.6\% &\bf+3.7\% &\bf-25.1\% &\bf+1.9\% &\bf+7.8\% &\bf-2.4\%\\
    \hline 
    \hline 
    Full-shot &99.1($\pm$0.2) &99.5($\pm$0.2) &97.5($\pm$0.3) &99.8($\pm$0.0) &96.7($\pm$0.5) &98.5 \\
    Ours      &98.3($\pm$0.5) &99.4($\pm$0.1) &97.3($\pm$0.2) &99.9($\pm$0.1) &95.6($\pm$0.3) &\bf{98.1} \\
    \hline
    \textbf{Gain(+\%)}  &\bf-0.8\% &\bf-0.1\% &\bf-0.2\% &\bf+0.1\% &\bf-1.1\% &\bf-0.4\%\\
    \hline 
    \hline   
\end{tabular}
}
\label{tb:Generalization test on MVTEC-AD(Efficient-AD,All classes(texture type)}
\end{table*}

\begin{table*}[ht!]
\caption{\textbf{Generalization test on BTAD(Patchcore, All classes, Detection AUROC)},
As a generalization test for BTAD dataset~\cite{ref:btad}, all class values of the patchcore~\cite{ref:patchcore} baseline and their average values were expressed. In the case of Class 1 and Class 3, it shows the object form, and in the case of Class 2, it shows the texture form.
All experimental results were measured using the official code provided by the author for each baseline and calculated as the average value performed five times.
}
  \centering
\centering
\resizebox{.5\linewidth}{!}{
\begin{tabular}{l|ccc|c}
    \noalign{\smallskip}\noalign{\smallskip}\hline\hline
                    \bf{}& Class-1& Class-2& Class-3& Avg \\
    \hline
    One-shot &72.2($\pm$7.2) &73.3($\pm$1.7) &64.6($\pm$1.7) &70.0 \\
    Ours     &92.4($\pm$0.3) &77.3($\pm$0.4) &70.5($\pm$0.8) &\bf80.1 \\
    \hline  
    \textbf{Gain(+\%)}  &\bf+20.2\% &\bf+4.0\% &\bf+5.9\% &\bf+10.1\% \\
    \hline 
    \hline   
    Few-shot &91.7($\pm$0.7) &80.5($\pm$0.7) &67.7($\pm$2.4) &80.0 \\
    Ours     &94.4($\pm$0.3) &80.7($\pm$0.8) &68.3($\pm$1.0) &\bf81.1 \\
    \hline
    \textbf{Gain(+\%)}  &\bf+2.7\% &\bf+0.2\% &\bf+0.6\% &\bf+1.1\% \\
    \hline 
    \hline   
    Full-shot &94.3($\pm$0.5) &81.8($\pm$0.6) &68.6($\pm$1.1) &81.6 \\
    Ours      &94.1($\pm$0.9) &82.4($\pm$0.7) &67.5($\pm$0.5) &\bf{81.3} \\
    \hline
    \textbf{Gain(+\%)}  &\bf-0.2\% &\bf+0.6\% &\bf-1.1\% &\bf-0.3\% \\
    \hline 
    \hline   
\end{tabular}
}
\label{tb:Generalization test on BTAD(Patchcore,All classes)}
\end{table*}

\begin{figure*}[t]
\centering
    \includegraphics[width=\linewidth]{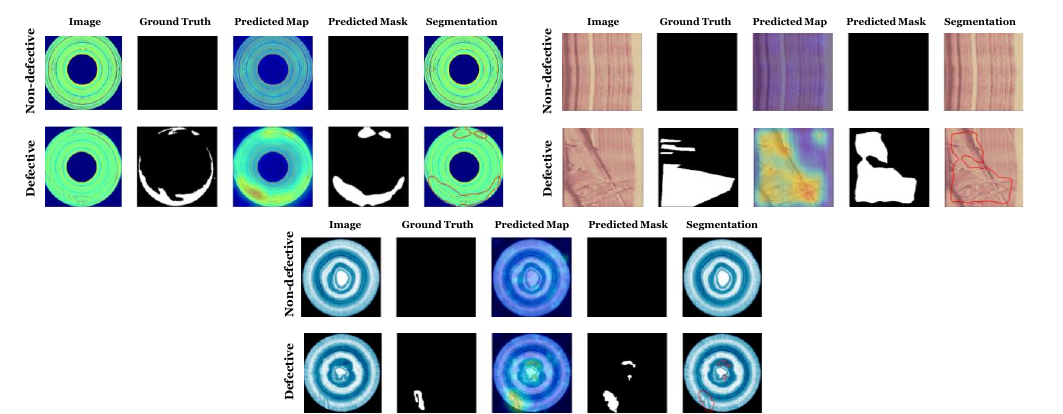}
  \caption{\textbf{Qualitative results[All classes] on BTAD dataset}. As shown in the figure, our model obtained qualitative results between non-defective and defect images for the three classes.
} 
  \label{fig:Segmentation results[All] on BTAD dataset}
\end{figure*}

\begin{table*}[ht!]
\caption{\textbf{Generalization test on MVTec-LOCO AD(Patchcore, All classes, Detection AUROC)},
As a generalization test for  MVTec-LOCO AD dataset~\cite{ref:mvtec_loco}, all class values of the patchcore~\cite{ref:patchcore} baseline, and their average values include logical anomalies and structural anomalies for each class were expressed.
All experimental results were measured using the official code provided by the author for each baseline and calculated as the average value performed five times.
}
  \centering
\centering
\resizebox{.7\linewidth}{!}{
\begin{tabular}{l|ccccc|c}
    \noalign{\smallskip}\noalign{\smallskip}\hline\hline
                    \bf{} &Class-1 &Class-2 &Class-3 &Class-4 &Class-5 &Avg \\
    \hline
    One-shot &59.4($\pm$0.2) &60.6($\pm$11.9) &54.1($\pm$2.7) &43.0($\pm$0.5) &62.1($\pm$1.4) &55.8 \\
    Ours     &74.9($\pm$1.4) &71.2($\pm$6.0) &59.1($\pm$0.5) &46.6($\pm$0.3) &64.0($\pm$0.4) &\bf63.2 \\
    \hline    
    \textbf{Gain(+\%)}  &\bf+15.5\% &\bf+10.6\% &\bf+5.0\% &\bf+3.6\% &\bf+1.9\% &\bf+7.4\% \\
    \hline 
    \hline   
    Few-shot &62.3($\pm$1.4) &83.4($\pm$0.8) &54.4($\pm$2.2) &54.6($\pm$0.7) &65.7($\pm$0.5) &64.1 \\
    Ours     &67.0($\pm$1.3) &91.4($\pm$0.4) &54.8($\pm$0.7) &53.5($\pm$1.2) &70.5($\pm$0.3) &\bf67.4 \\
    \hline
    \textbf{Gain(+\%)}  &\bf+4.7\% &\bf+8.0\% &\bf+0.4\% &\bf-1.1\% &\bf+4.8\% &\bf+3.3\% \\
    \hline 
    \hline     
    Full-shot &79.7($\pm$0.6) &96.0($\pm$0.3) &74.0($\pm$1.0) &62.5($\pm$0.9) &81.8($\pm$0.4) &78.8 \\
    Ours  &78.7($\pm$1.0) &96.3($\pm$0.2) &72.4($\pm$1.1) &63.7($\pm$1.1) &81.6($\pm$1.0) &\bf{78.5} \\
    \hline
    \textbf{Gain(+\%)}  &\bf-1.0\% &\bf+0.3\% &\bf-1.6\% &\bf+1.2\% &\bf-0.2\% &\bf-0.3\% \\
    \hline 
    \hline 
\end{tabular}
}
\label{tb:Generalization test on MVTEC-LOCO(Patchcore,All classes)}
\end{table*}

\subsection{Detailed Experimental Result}

\subsubsection{Comprehensive Comparison}
In Table~\ref{tb:Performance increase rate by total percentage used original dataset}, our method shows improved performance compared to all baselines.
In particular, good performance can be obtained even under limited conditions (one- or few-shot scenarios), indicating that the model effectively incorporates features of non-defective images into the images generated by our method.

\subsubsection{Additional Quantitative Results}
We show our performance against each baseline, organized across all scenarios (one, few, full-shot) and all classes.
First, we compared the average performance by dividing the object type (Tables~\ref{tb:Generalization test on MVTEC-AD(Pathcore,All classes(object type)},~\ref{tb:Generalization test on MVTEC-AD(Cflow,All classes(object type)},~\ref{tb:Generalization test on MVTEC-AD(Reverse-distillation,All classes(object type)},~\ref{tb:Generalization test on MVTEC-AD(Efficient-AD,All classes(object type)}) and texture types (Tables~\ref{tb:Generalization test on MVTEC-AD(Pathcore,All classes(texture type)},~\ref{tb:Generalization test on MVTEC-AD(Cflow,All classes(texture type)},~\ref{tb:Generalization test on MVTEC-AD(Reverse-distillation,All classes(texture type)},~\ref{tb:Generalization test on MVTEC-AD(Efficient-AD,All classes(texture type)}).
In the Patchcore, Cflow, and Efficient-AD models, the Object Type shows significantly higher performance, but in the reverse distillation model, the Texture Type shows higher performance.

Therefore, we revalidated it on the BTAD dataset (Table~\ref{tb:Generalization test on BTAD(Patchcore,All classes)}) in which object type and texture type, and it shows the highest performance in object type.
Finally, we tested it on the complex MVTec-LOCO AD dataset (Table~\ref{tb:Generalization test on MVTEC-LOCO(Patchcore,All classes)}) and showed performance improvement overall. 
In particular, the one-shot scenario showed good performance, improving by 15.5\% and 10.6\% in the breakfast box and juice bottle classes.

\begin{figure}[t]
\centering
    \includegraphics[width=\linewidth]{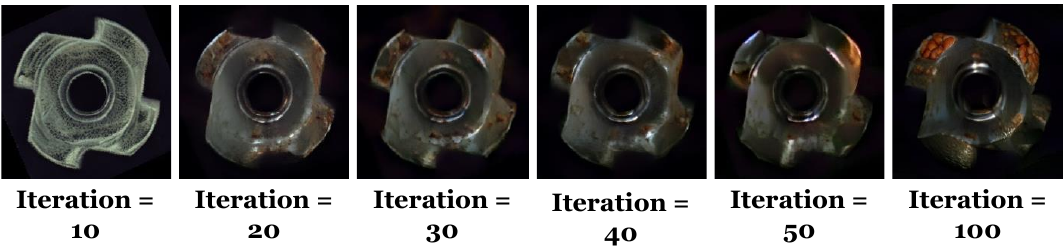}
    \includegraphics[width=.6\linewidth]{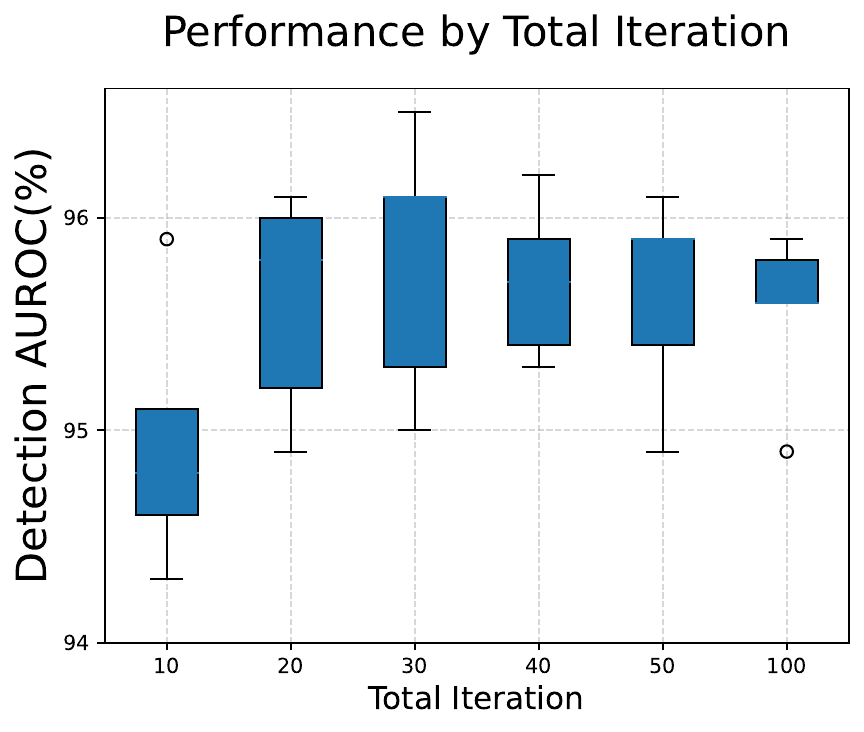}
    \vspace{-0.3cm}
\caption{
\textbf{Comparison results of images generated according to the number of iterations.} (Reverse-distillation, Metal-nut(class)),
The generated images(top) show that strongly reflects input text information as repetition increases.
The performance results(bottom) show a trend in which performance variance gradually decreases with saturation at a certain point.}
  \label{fig:Performance trend Results by Iteration}
    \vspace{-0.2cm}
\end{figure}

\subsubsection{Additional Qualitative Results}
We show the qualitative results for representative classes, including the test images with ground truth masks for detection and the anomaly localization score heatmap for segmentation. 
These results can be found in Figs.~\ref{fig:Detection and Segmentation results[All] on MVTecAD dataset},~\ref{fig:Segmentation results[All] on BTAD dataset} for the MVTecAD and BTAD datasets, respectively.

\begin{table}[t]
\caption{
\textbf{Ablation studies by Variance-aware parameters.}
The result of comparing evaluation metrics (SSIM, PSNR, VIF, LPIPS) that measure the similarity and quality of the original image and a single generated image. 
The image generated by each parameter of the latent vector distribution in the variance-aware method was compared with the original image.} 
  \centering
\centering
    \vspace{-0.4cm}
\resizebox{\linewidth}{!}{
\begin{tabular}{l|c|c|c|c|c}
    \noalign{\smallskip}\noalign{\smallskip}\hline\hline
            Method &Avg & SSIM($\uparrow$) & PSNR($\uparrow$) & VIF($\uparrow$) & LPIPS($\downarrow$)\\
    \hline
    \hline   
    \text (a) Original                              &87.7   &1.00 &- &1.00   &0.00 \\ 
    \hline
    \text (b) $\mu$                                &87.8  &0.88 &26.60 &0.09   &\bf664.31 \\ 
    \text (c) $\mu$ + $\sigma$                     &88.4  &0.72 &22.23 &0.09   &784.34 \\  
    \text (d) $\mu$ + ($\sigma$ x $\epsilon$)      &\bf88.6  &\bf0.88 &\bf26.83 &\bf0.11   &726.94 \\  
    \hline
    \hline
\end{tabular}
}
    \vspace{-0.4cm}
\label{tb:Comparison results by parameter}
\end{table}

\begin{figure}[t]
\centering
    \includegraphics[width=\linewidth]{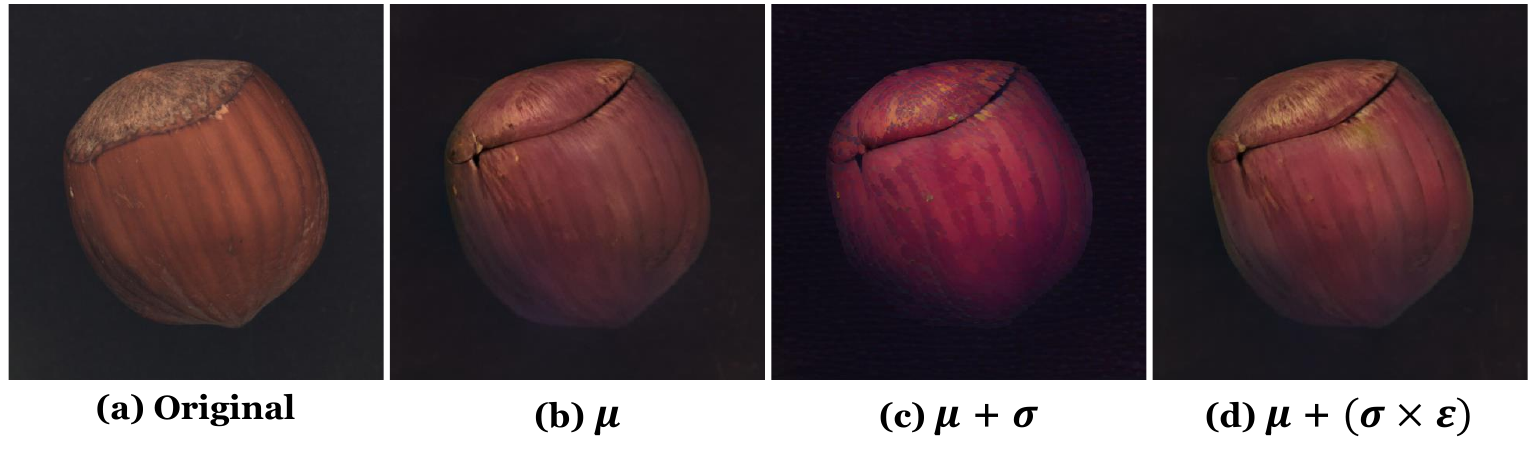}
    \vspace{-0.8cm}
\caption{
\textbf{Qualitative results by Variance-aware method parameters.}}
  \label{fig:Qualitative Results by parameters}
\end{figure}

\subsection{Additional Analysis Results}

\subsubsection{Visual analysis on t-SNE distributions}
Fig.~\ref{fig:t-sne results} shows the T-SNE distributions to compare the latent features of original non-defective images in blue dots, our generated images in red dots, and defective images in green dots.
The results show that the samples generated by our model are evenly distributed close to the non-defective data while effectively separating the defective data.

\begin{figure*}[t]
\centering
    \includegraphics[width=\linewidth]{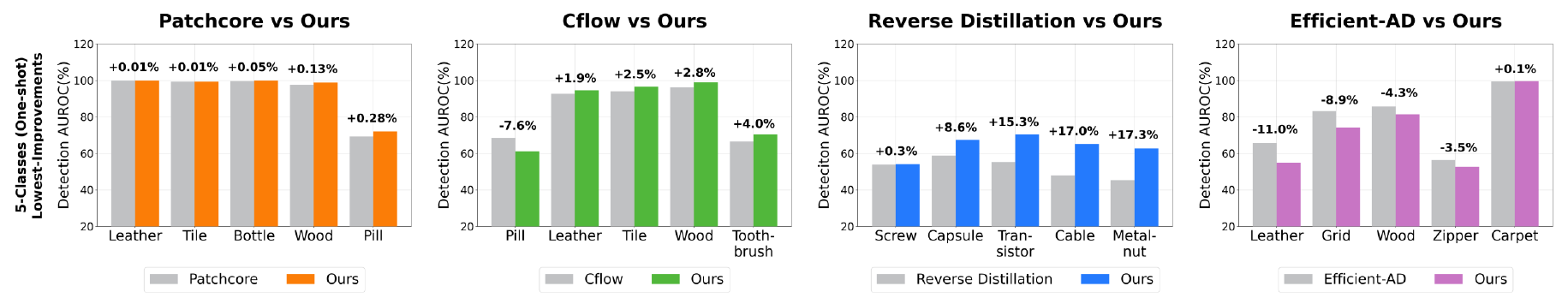}
  \caption{\textbf{Lowest-improving results for anomaly detection in MVTecAD dataset.}
  The results shows the average score for the lowest-improving five classes.}
    \vspace{-0.2cm}
  \label{fig:Generalization test of lowest anomaly detection on MVTec}
\end{figure*}

\subsubsection{Comparison Results of images generated according to the number of iterations}
The generated images depicted in Fig.~\ref{fig:Performance trend Results by Iteration}(top) show a notable increase in the reflection of input text (class name) information as the number of training iterations in the model increases. 
This reduces the learning obstacles of the original image and allows for a more strongly represented representation of its properties.
On the other hand, Fig.~\ref{fig:Performance trend Results by Iteration}(bottom) reveals that achieved the highest level of performance between the 20th and 30th iterations. During the initial iteration, the standard deviation value is approximately 1.57, but it rapidly decreases as the model iterates, reaching 0.39. 
Therefore, it is essential to integrate relevant textual information under appropriate training iteration conditions to obtain optimal results.

\subsubsection{Ablation studies by Variance-aware parameters}
As shown by Table~\ref{tb:Comparison results by parameter}, Fig.~\ref{fig:Qualitative Results by parameters}, we compared the image quality effects through an ablation test of the variance-aware parameters elements Mu, sigma, and epsilon values. 
The SSIM, which considers visual elements such as image structure, contrast, and texture, was found to be most structurally similar when we reflected all parameters.
The PSNR, which measures the difference between two images as the difference in pixel values, also shows the best score, which indicates good quality.
VIF(Visual Information Fidelity), used to evaluate the quality of images with compression or loss, also shows the best score.

\begin{table}[t]
\caption{
\textbf{Performance improvement experiment.} Based on the lowest improvement results collected from Fig.~\ref{fig:Generalization test of lowest anomaly detection on MVTec}, we attempt to improve the performance by changing the pre-trained CLIP model and augmentation strategy.
(Augmentation strategies : Strategies-1 : RandomCrop, ColorJitter / 
                              Strategies-2 : RandomRotation, RandomAutocontrast)}
\centering
\resizebox{\linewidth}{!}{
\begin{tabular}{l|c|c|c|c}
    \noalign{\smallskip}\noalign{\smallskip}\hline\hline
    Component            &Leather(Texture) &Gain(\%)&Pill(Object) &Gain(\%)\\
    \hline
    \hline
    Baseline             &65.8($\pm$0.0)    &-    &68.6($\pm$0.0)     &-      \\
    \hline
    \hline
    1) CLIP model        &                                       \\
    \hline
    ViT-B/16              &54.8($\pm$8.6) &(-11.0\%) &61.0($\pm$6.5) &(-7.6\%)\\
    ResNet50x64           &\bf76.1($\pm$0.1) &\bf(+10.3\%) &\bf71.5($\pm$0.0) &\bf(+2.9\%)\\
    \hline
    2) Augmentation                                                                  \\
    \hline
    Strategies-1          &65.8($\pm$0.0) &(+0.0\%) &68.6($\pm$0.0)  &(+0.0\%)\\
    Strategies-2          &56.1($\pm$0.0) &(-9.7\%) &58.8($\pm$0.0)   &(-9.8\%)\\
    \hline
    \hline
\end{tabular}
}
\label{tb:networks and the outcomes of those experiments}
\end{table}

\begin{table}[t]
\caption{\textbf{Comparison with Data Augmentation Strategies.}
A one-shot scenario experiment was conducted on a toothbrush (Object type) and grid class (Texture type), and performance was compared to the baseline using Patchcore.
}
\centering
\resizebox{\linewidth}{!}{
\begin{tabular}{l|c|c|c|c}
    \noalign{\smallskip}\noalign{\smallskip}\hline\hline
    Component            &Toothbrush(Object) &Gain(\%)&Grid(Texture) &Gain(\%)\\
    \hline
    \hline
    Baseline                                        &74.7   &-               &68.6    &-      \\
    \hline  
    Strategies-1                                    &76.9 &(+2.2\%)          &61.8  &(-6.8\%)\\
    Strategies-2                                    &77.1 &(+2.4\%)          &62.2  &(-6.4\%)\\
    AutoAugmentation ~\cite{ref:aa}                 &74.4 &(-0.3\%)          &67.8  &(-0.8\%)\\
    RandAugmentation ~\cite{ref:ra}                 &76.7 &(+2.0\%)          &70.3  &(+1.7\%)\\
    \hline
    \bf{Ours}                                       &\bf78.9 &\bf(+4.2\%)    &\bf72.2  &\bf(+3.6\%)\\
    \hline
    \hline
\end{tabular}
}
\label{tb:Ablataion test by DA}
\end{table}

\subsection{Discussion}

\subsubsection{Lowest-improving results for anomaly detection in MVTecAD dataset.}
Additionally, we find scenarios where performance was low and analyze how to improve them.
In Fig.~\ref{fig:Generalization test of lowest anomaly detection on MVTec}, the targets with the most significant decrease in performance were the Leather class in the 4th graph and the Pill class in the 2nd graph.
As shown by Table~\ref{tb:networks and the outcomes of those experiments}, when we changed the CLIP model to ResNet50x64, the Leather class(texture type) was significantly improved to about 10.1\%, and the variation also tended to change stably. 
However, when the augmentation strategy changed, it was ineffective in performance.
Therefore, we analyzed the effectiveness of the augmentation strategy through additional experiments in Table~\ref{tb:Ablataion test by DA}.

\begin{figure}[t]
\centering
    \includegraphics[width=\linewidth]{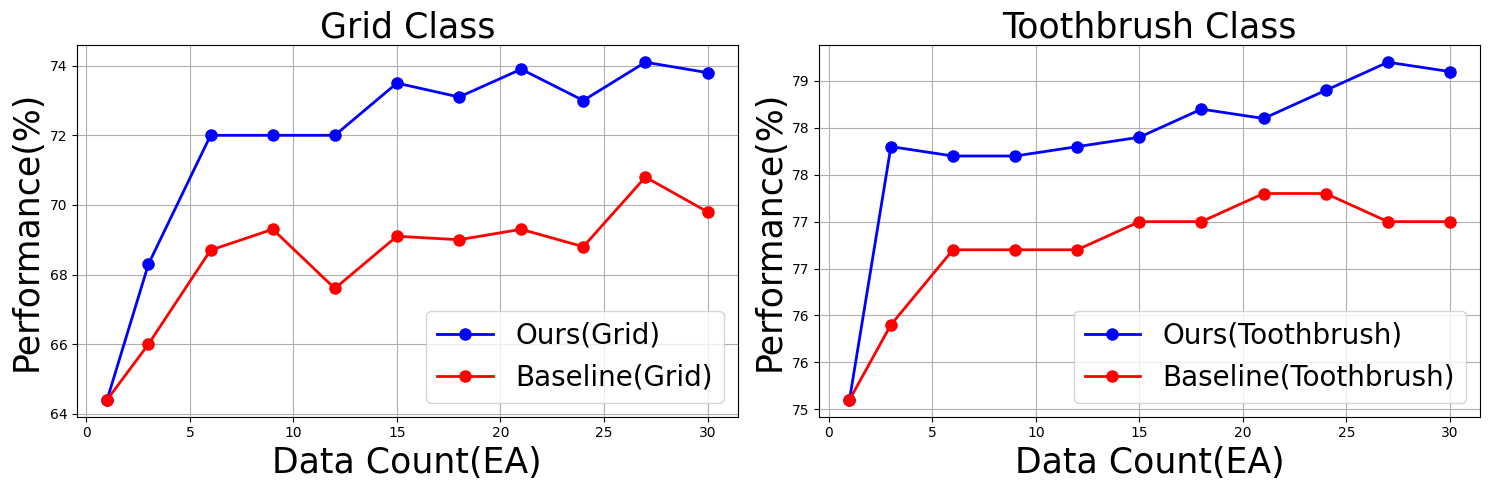}
    \vspace{-0.6cm}
\caption{
\textbf{Necessity of generating Non-defective data.} 
The addition of non-defective data(non-various images in Baseline) suffers from the rapid convergence of performance. 
However, through our generative model, a variety of non-defective data can be acquired to improve the performance, even with few real non-defective images. 
}
    \vspace{-0.3cm}
  \label{fig:Necessity of Generating Non-defective Data}
\end{figure}

\subsubsection{Various augmentation strategies.}
In Table~\ref{tb:Ablataion test by DA}, the possibility of improvement was analyzed by applying various augmentation strategies when generating non-defective images.
Effective strategies differ depending on object type and texture type.
For example, Strategies 1,2 presented were effective for object types such as Toothbrushes but not for texture types such as Grids.
Therefore, we experimented with a random selection of Gaussian Blur, NoiseReduction, RandomRotation, RandomAdjustSharpness, RandomAutocontrast, and ColorJitter strategies that we empirically found to be effective. As a result, we confirmed the possibility of improving performance when specific strategies are used appropriately.

\subsubsection{Design with connection to anomaly detection and segmentation.}
Additional non-defective images do not necessarily increase the performance due to their duplicity, while the images generated by our approach show more effectiveness than the same number of real images. 
Our method is designed to generate non-defective data, preserving the possible variance of input data through text-based guidance. 
This approach is particularly relevant in industrial anomaly detection and segmentation, where outlier scores are based on the distance between test and non-defective images.
The importance of non-defective generation is also validated through the experiments of Fig.~\ref{fig:Necessity of Generating Non-defective Data}.

\end{document}

%% file: preamble.tex
\usepackage[dvipsnames]{xcolor}


%% file: main.bbl
\begin{thebibliography}{30}
\providecommand{\natexlab}[1]{#1}
\providecommand{\url}[1]{\texttt{#1}}
\expandafter\ifx\csname urlstyle\endcsname\relax
  \providecommand{\doi}[1]{doi: #1}\else
  \providecommand{\doi}{doi: \begingroup \urlstyle{rm}\Url}\fi

\bibitem[Bae et~al.(2022)Bae, Lee, and Kim]{ref:PNI}
Jaehyeok Bae, Jae-Han Lee, and Seyun Kim.
\newblock Image anomaly detection and localization with position and neighborhood information.
\newblock \emph{arXiv}, 2022.

\bibitem[Batzner et~al.(2023)Batzner, Heckler, and K{\"o}nig]{ref:efficientad}
Kilian Batzner, Lars Heckler, and Rebecca K{\"o}nig.
\newblock Efficientad: Accurate visual anomaly detection at millisecond-level latencies.
\newblock \emph{arXiv}, 2023.

\bibitem[Bergmann et~al.(2019)Bergmann, Fauser, Sattlegger, and Steger]{ref:mvtec}
Paul Bergmann, Michael Fauser, David Sattlegger, and Carsten Steger.
\newblock Mvtec ad--a comprehensive real-world dataset for unsupervised anomaly detection.
\newblock In \emph{Proceedings of the IEEE/CVF conference on computer vision and pattern recognition}, pages 9592--9600, 2019.

\bibitem[Bergmann et~al.(2022)Bergmann, Batzner, Fauser, Sattlegger, and Steger]{ref:mvtec_loco}
Paul Bergmann, Kilian Batzner, Michael Fauser, David Sattlegger, and Carsten Steger.
\newblock Beyond dents and scratches: Logical constraints in unsupervised anomaly detection and localization.
\newblock \emph{International Journal of Computer Vision}, 130\penalty0 (4):\penalty0 947--969, 2022.

\bibitem[Crowson et~al.(2022)Crowson, Biderman, Kornis, Stander, Hallahan, Castricato, and Raff]{ref:vqgan-clip}
Katherine Crowson, Stella Biderman, Daniel Kornis, Dashiell Stander, Eric Hallahan, Louis Castricato, and Edward Raff.
\newblock Vqgan-clip: Open domain image generation and editing with natural language guidance.
\newblock In \emph{Computer Vision--ECCV 2022}, pages 88--105. Springer, 2022.

\bibitem[Cubuk et~al.(2019)Cubuk, Zoph, Mane, Vasudevan, and Le]{ref:aa}
Ekin~D Cubuk, Barret Zoph, Dandelion Mane, Vijay Vasudevan, and Quoc~V Le.
\newblock Autoaugment: Learning augmentation strategies from data.
\newblock In \emph{Proceedings of the IEEE/CVF Conference on Computer Vision and Pattern Recognition}, pages 113--123, 2019.

\bibitem[Cubuk et~al.(2020)Cubuk, Zoph, Shlens, and Le]{ref:ra}
Ekin~D Cubuk, Barret Zoph, Jonathon Shlens, and Quoc~V Le.
\newblock Randaugment: Practical automated data augmentation with a reduced search space.
\newblock In \emph{Proceedings of the IEEE/CVF Conference on Computer Vision and Pattern Recognition Workshops}, pages 702--703, 2020.

\bibitem[Defard et~al.(2021)Defard, Setkov, Loesch, and Audigier]{ref:padim}
Thomas Defard, Aleksandr Setkov, Angelique Loesch, and Romaric Audigier.
\newblock Padim: a patch distribution modeling framework for anomaly detection and localization.
\newblock In \emph{Pattern Recognition. ICPR International Workshops and Challenges: Virtual Event, January 10--15, 2021, Proceedings, Part IV}, pages 475--489. Springer, 2021.

\bibitem[Deng and Li(2022)]{ref:Reverse_Distillation}
Hanqiu Deng and Xingyu Li.
\newblock Anomaly detection via reverse distillation from one-class embedding.
\newblock In \emph{Proceedings of the IEEE/CVF Conference on Computer Vision and Pattern Recognition}, pages 9737--9746, 2022.

\bibitem[Esser et~al.(2020)Esser, Rombach, and Ommer]{ref:vqgan}
Patrick Esser, Robin Rombach, and Björn Ommer.
\newblock Taming transformers for high-resolution image synthesis, 2020.

\bibitem[Fellbaum(2005)]{ref:wordnet}
Christiane Fellbaum.
\newblock Wordnet and wordnets.
\newblock In \emph{Encyclopedia of Language and Linguistics}, pages 665--670. Elsevier, 2005.

\bibitem[Goodfellow et~al.(2014)Goodfellow, Pouget-Abadie, Mirza, Xu, Warde-Farley, Ozair, Courville, and Bengio]{ref:gan}
Ian Goodfellow, Jean Pouget-Abadie, Mehdi Mirza, Bing Xu, David Warde-Farley, Sherjil Ozair, Aaron Courville, and Yoshua Bengio.
\newblock Generative adversarial nets.
\newblock \emph{Advances in neural information processing systems}, 27, 2014.

\bibitem[Google(2023)]{ref:GoogleImages}
Google.
\newblock Google image search.
\newblock \url{https://www.google.com/imghp?hl=ko&ogbl}, 2023.
\newblock Accessed on 2023.

\bibitem[Gudovskiy et~al.(2022)Gudovskiy, Ishizaka, and Kozuka]{ref:cflow}
Denis Gudovskiy, Shun Ishizaka, and Kazuki Kozuka.
\newblock Cflow-ad: Real-time unsupervised anomaly detection with localization via conditional normalizing flows.
\newblock In \emph{Proceedings of the IEEE/CVF Winter Conference on Applications of Computer Vision}, pages 98--107, 2022.

\bibitem[Jeong et~al.(2023)Jeong, Zou, Kim, Zhang, Ravichandran, and Dabeer]{ref:winclip}
Jongheon Jeong, Yang Zou, Taewan Kim, Dongqing Zhang, Avinash Ravichandran, and Onkar Dabeer.
\newblock Winclip: Zero-/few-shot anomaly classification and segmentation.
\newblock In \emph{Proceedings of the IEEE/CVF Conference on Computer Vision and Pattern Recognition}, pages 19606--19616, 2023.

\bibitem[Li et~al.(2022)Li, Xu, Wang, Zhou, Lin, Zhu, Zeng, Ji, and Chang]{ref:cliphome}
Manling Li, Ruochen Xu, Shuohang Wang, Luowei Zhou, Xudong Lin, Chenguang Zhu, Michael Zeng, Heng Ji, and Shih-Fu Chang.
\newblock Clip-event: Connecting text and images with event structures.
\newblock In \emph{Proceedings of the IEEE/CVF Conference on Computer Vision and Pattern Recognition}, pages 16420--16429, 2022.

\bibitem[MidJourney(2023)]{ref:MidJourney}
MidJourney.
\newblock Midjourney home page.
\newblock \url{https://www.midjourney.com/home}, 2023.
\newblock Accessed on 2023.

\bibitem[Mishra et~al.(2021)Mishra, Verk, Fornasier, Piciarelli, and Foresti]{ref:btad}
Pankaj Mishra, Riccardo Verk, Daniele Fornasier, Claudio Piciarelli, and Gian~Luca Foresti.
\newblock Vt-adl: A vision transformer network for image anomaly detection and localization.
\newblock In \emph{2021 IEEE 30th International Symposium on Industrial Electronics}, pages 01--06. IEEE, 2021.

\bibitem[Radford et~al.(2021)Radford, Kim, Hallacy, Ramesh, Goh, Agarwal, Sastry, Askell, Mishkin, Clark, et~al.]{ref:clip}
Alec Radford, Jong~Wook Kim, Chris Hallacy, Aditya Ramesh, Gabriel Goh, Sandhini Agarwal, Girish Sastry, Amanda Askell, Pamela Mishkin, Jack Clark, et~al.
\newblock Learning transferable visual models from natural language supervision.
\newblock In \emph{International conference on machine learning}, pages 8748--8763. Proceedings of Machine Learning Research, 2021.

\bibitem[Ramesh et~al.(2021)Ramesh, Pavlov, Goh, Gray, Voss, Radford, Chen, and Sutskever]{ref:dall-e}
Aditya Ramesh, Mikhail Pavlov, Gabriel Goh, Scott Gray, Chelsea Voss, Alec Radford, Mark Chen, and Ilya Sutskever.
\newblock Zero-shot text-to-image generation.
\newblock In \emph{International Conference on Machine Learning}, pages 8821--8831. PMLR, 2021.

\bibitem[Roth et~al.(2021)Roth, Pemula, Zepeda, Schölkopf, Brox, and Gehler]{ref:patchcore}
Karsten Roth, Latha Pemula, Joaquin Zepeda, Bernhard Schölkopf, Thomas Brox, and Peter Gehler.
\newblock Towards total recall in industrial anomaly detection, 2021.

\bibitem[Rudolph et~al.(2021)Rudolph, Wandt, and Rosenhahn]{ref:differnet}
Marco Rudolph, Bastian Wandt, and Bodo Rosenhahn.
\newblock Same same but differnet: Semi-supervised defect detection with normalizing flows.
\newblock In \emph{Proceedings of the IEEE/CVF winter conference on applications of computer vision}, pages 1907--1916, 2021.

\bibitem[Rudolph et~al.(2022)Rudolph, Wehrbein, Rosenhahn, and Wandt]{ref:cross-scale-flows}
Marco Rudolph, Tom Wehrbein, Bodo Rosenhahn, and Bastian Wandt.
\newblock Fully convolutional cross-scale-flows for image-based defect detection.
\newblock In \emph{Proceedings of the IEEE/CVF Winter Conference on Applications of Computer Vision}, pages 1088--1097, 2022.

\bibitem[Ruff et~al.(2018)Ruff, Vandermeulen, Goernitz, Deecke, Siddiqui, Binder, M{\"u}ller, and Kloft]{ref:svdd}
Lukas Ruff, Robert Vandermeulen, Nico Goernitz, Lucas Deecke, Shoaib~Ahmed Siddiqui, Alexander Binder, Emmanuel M{\"u}ller, and Marius Kloft.
\newblock Deep one-class classification.
\newblock In \emph{International conference on machine learning}, pages 4393--4402. Proceedings of Machine Learning Research, 2018.

\bibitem[Ruff et~al.(2019)Ruff, Vandermeulen, G{\"o}rnitz, Binder, M{\"u}ller, M{\"u}ller, and Kloft]{ref:deep-svdd}
Lukas Ruff, Robert~A Vandermeulen, Nico G{\"o}rnitz, Alexander Binder, Emmanuel M{\"u}ller, Klaus-Robert M{\"u}ller, and Marius Kloft.
\newblock Deep semi-supervised anomaly detection.
\newblock \emph{arXiv}, 2019.

\bibitem[Sheynin et~al.(2021)Sheynin, Benaim, and Wolf]{ref:fewanomaly}
Shelly Sheynin, Sagie Benaim, and Lior Wolf.
\newblock A hierarchical transformation-discriminating generative model for few shot anomaly detection.
\newblock In \emph{Proceedings of the IEEE/CVF International Conference on Computer Vision}, pages 8495--8504, 2021.

\bibitem[Yang et~al.(2023)Yang, Wu, and Feng]{ref:memseg}
Minghui Yang, Peng Wu, and Hui Feng.
\newblock Memseg: A semi-supervised method for image surface defect detection using differences and commonalities.
\newblock \emph{Engineering Applications of Artificial Intelligence}, 119:\penalty0 105835, 2023.

\bibitem[Yi and Yoon(2020)]{ref:patch-svdd}
Jihun Yi and Sungroh Yoon.
\newblock Patch svdd: Patch-level svdd for anomaly detection and segmentation.
\newblock In \emph{Proceedings of the Asian Conference on Computer Vision}, 2020.

\bibitem[Yu et~al.(2021)Yu, Zheng, Wang, Li, Wu, Zhao, and Wu]{ref:fastflow}
Jiawei Yu, Ye Zheng, Xiang Wang, Wei Li, Yushuang Wu, Rui Zhao, and Liwei Wu.
\newblock Fastflow: Unsupervised anomaly detection and localization via 2d normalizing flows.
\newblock \emph{arXiv}, 2021.

\bibitem[Zavrtanik et~al.(2021)Zavrtanik, Kristan, and Sko{\v{c}}aj]{ref:draem}
Vitjan Zavrtanik, Matej Kristan, and Danijel Sko{\v{c}}aj.
\newblock Draem-a discriminatively trained reconstruction embedding for surface anomaly detection.
\newblock In \emph{Proceedings of the IEEE/CVF International Conference on Computer Vision}, pages 8330--8339, 2021.

\end{thebibliography}
